\pdfoutput=1

\documentclass[11pt]{article}

\usepackage[]{acl}

\usepackage{times}
\usepackage{latexsym}

\usepackage{booktabs}
\usepackage{graphicx}
\usepackage{subfig}
\usepackage{colortbl}
\usepackage{tabularx}
\newcolumntype{Y}{>{\centering\arraybackslash}X}
\usepackage[textsize=tiny]{todonotes}
\usepackage{listings}
\usepackage[utf8]{inputenc}
\usepackage{tcolorbox}
\usepackage{enumitem}
\usepackage{multicol}
\usepackage{float}
\usepackage{placeins}

\usepackage[T1]{fontenc}

\usepackage[utf8]{inputenc}

\usepackage{microtype}

\usepackage{inconsolata}

\usepackage{graphicx}

\title{Can LLM be a Personalized Judge?}
\author{Yijiang River Dong$^*$ \and Tiancheng Hu$^*$ \and Nigel Collier \\ \{yd358, th656, nhc30\}@cam.ac.uk \\
         University of Cambridge}

\begin{document}
\maketitle
\def\thefootnote{*}\footnotetext{Equal contribution}\def\thefootnote{\arabic{footnote}}

\begin{abstract}
Ensuring that large language models (LLMs) reflect diverse user values and preferences is crucial as their user bases expand globally. It is therefore encouraging to see the growing interest in LLM personalization within the research community. However, current works often rely on the LLM-as-a-Judge approach for evaluation without thoroughly examining its validity. In this paper, we investigate the reliability of LLM-as-a-\textbf{Personalized}-Judge—asking LLMs to judge user preferences based on personas. Our findings suggest that directly applying LLM-as-a-Personalized-Judge is less reliable than previously assumed, showing low and inconsistent agreement with human ground truth. The personas typically used are often overly simplistic, resulting in low predictive power. To address these issues, we introduce verbal uncertainty estimation into the LLM-as-a-Personalized-Judge pipeline, allowing the model to express low confidence on uncertain judgments. This adjustment leads to much higher agreement (above 80\%) on high-certainty samples for binary tasks.  Through human evaluation, we find that the LLM-as-a-Personalized-Judge achieves comparable performance to third-party humans evaluation and even surpasses human performance on high-certainty samples. Our work indicates that certainty-enhanced LLM-as-a-Personalized-Judge offers a promising direction for developing more reliable and scalable methods for evaluating LLM personalization. Our code is available at \url{https://github.com/dong-river/Personalized-Judge}.
\end{abstract}

\section{Introduction}
\begin{figure*}[htbp!]
\centering
\vspace{-100pt} %
\includegraphics[trim=0 30 0 30, clip, scale=0.57]{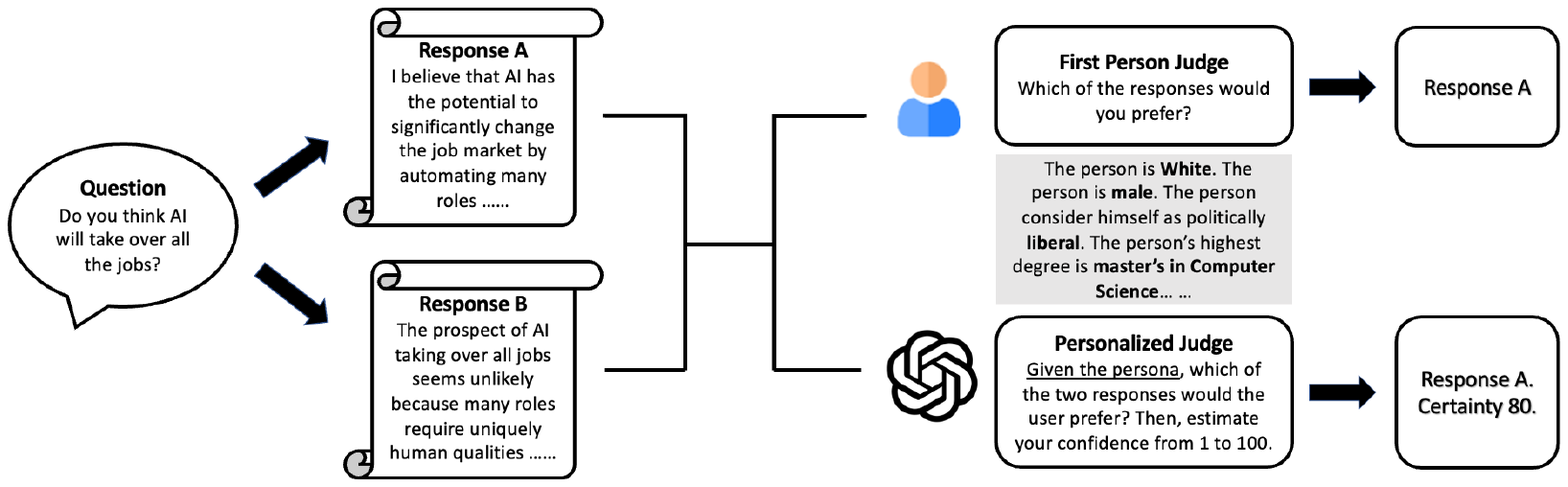}
\vspace{-90pt} %
\caption{\textbf{Overall workflow of Personalized Judge.} Given a subjective question and two distinct responses, we ask an LLM to infer the preference of a real user based on a user persona. We also ask the LLM to estimate its certainty level in this prediction. The inferred preference is then compared against the user's self-reported ground truth to evaluate the performance of the Judge.} 
\label{workflow}
\end{figure*}

As large language models (LLMs) gain widespread adoption among global users with diverse backgrounds, it is imperative to ensure these models designed to reflect their values and preferences~\cite{sorensen_roadmap_2024, kirk2024prism}. However, the current alignment process often assumes a homogeneous set of human preferences and ignores individual perspectives, even in context-dependent, subjective tasks~\cite{santurkar2023whose}. Therefore, efforts have been made to fine-tune LLMs to encode individual preferences or enhance role-playing capabilities~\cite{jang2023personalized, shao_character-llm_2023, occhipinti2023prodigy, li2024culturellm, andukuri_star-gate_2024} with ``LLM-as-a-Judge'' as the main evaluation metric \cite{zheng2023judging}, often without adequate validation.

Despite ``LLM-as-a-Judge'' showing high agreement with human annotators in many tasks, its effectiveness for personalization tasks remains largely unscrutinized. MT-Bench~\cite{zheng2023judging} includes a role-playing component but only considered simplistic personas, such as "imagine you are a doctor," without addressing more complex personas that encompass demographics, user descriptions, and prior interactions—settings increasingly employed in recent research. Furthermore, a persona description may not always be contextually relevant. Knowing that someone is a doctor, for instance, provides little insight into their favorite types of beverages. We refer to this issue as the \textit{persona sparsity issue}.\footnote{Our use of the term ``persona sparsity'' diverges from works like~\citet{Zheng_Zhang_Huang_Mao_2020, song-etal-2021-bob}. While they typically refer to the scarcity of naturalistic dialog data directly reflecting persona variables, we highlight a related but distinct problem: the available persona variables may not offer an informed prior about the person involved for a specific task.}

In this paper, we examine the validity of LLM-as-a-Judge for personalization, where the objective is to generate personalized outputs based on a given user persona (see Figure~\ref{workflow}). We assess performance on tasks where ground truth data is available, including PRISM \cite{kirk2024prism}, OpinionQA \cite{santurkar2023whose}, Public Reddit \cite{staab2023beyond}, and Empathetic Conversation \cite{omitaomu2022empathic}. To address the issue of persona sparsity, we then propose a verbal uncertainty estimation component into the Judge pipeline. By articulating its own certainty levels, an LLM can assign lower certainty to samples for which it perceives insufficient predictive power. Additionally, we conduct a crowdsourcing experiment and compare the performance of LLM-as-a-Personalized-Judge to third-person human evaluation.

Our findings are as follows: (1) Contrary to previous assumptions, standard LLM-as-a-Judge is not sufficiently reliable for personalization tasks, showing only around 70\% agreement with human judgments in binary choice scenarios, and dropping below 60\% for certain tasks. (2) We identify persona sparsity as a major factor contributing to this unreliability. To address this, we introduce verbal uncertainty estimation into the LLM-as-Personalized-Judge process and achieve above 80\% performance in high-certainty samples. (3) In a crowdsourcing experiment, we find that LLM-as-a-Personalized-Judge achieves performance comparable to third-person\footnote{Here, first-person evaluation refers to judgments made by the individuals for whom the personalization is intended, reflecting their own preferences and values. Third-person evaluation involves external annotators who assess the personalization based on persona descriptions rather than personal preferences. } human judgment and even surpasses human performance on high-certainty samples. While first-person human evaluation from diverse backgrounds remains the gold standard for personalization, in the absence of such annotations, LLM-as-a-Personalized-Judge with certainty thresholding could serve as an effective and scalable alternative.

\section{Background and Related Work}

\label{sec:related_work}
\textbf{Personalization} in machine learning  refers to the process of tailoring a model's output to suit the unique preferences, needs, and behaviors of individual users (see~\citet{doi:10.1080/10919392.2006.9681199} for an in-depth discussion). This concept is at the core of recommender systems~\cite{10.1145/371920.372071}, and been explored in various contexts in NLP, such as dialogue system \cite{li-etal-2016-persona,zhang2018personalizing}, summarization~\cite{DIAZ20071715,yan-etal-2011-summarize}, user profiling and computational sociolinguistics~\cite{10.1162/COLI_a_00258}. These studies typically aim to understand the diverse linguistic patterns of users from varying backgrounds and contexts and to integrate persona information to enhance task performance. For surveys on these topics, see~\citet{flek-2020-returning, hovy-yang-2021-importance, yang2024call}.

In the context of LLMs, personalization has become even more critical due to the vast, diverse, and ever-growing user base. The necessity to align LLMs to a pluralistic set of user needs is discussed in \cite{sorensen_roadmap_2024}. However, the current alignment processes typically assume a single set of human preferences and researchers are just beginning to explore methods to address  the varied preferences and values of different users, either at the collective level~\cite{conitzer2024social,klingefjord2024human} or at individual level~\cite{salemi_lamp_2023,gao2024aligning,li_personalized_2024,jang2023personalized,wang_learning_2023}. Our study focuses on the evaluation aspect of personalized alignment approaches.

A challenging issue in this domain is the definition of personas. Not all variables are universally applicable or useful~\cite{hu_quantifying_2024}. For instance, while knowing an individual's profession as a doctor may offer some insights about this individual, it does not necessarily inform us about their preferred types of beverages. Ideally, we would include demographic, behavioral, and contextual factors that are relevant to the specific task at hand. However, defining the relevant set of variables a priori is inherently difficult. Additionally, even if surveys are designed to gather this information, acquiring such detailed information on a large scale is often impractical and can frequently result in incomplete responses. We refer to this challenge as the \textit{persona sparsity} issue. In practice, this means that in some cases, we cannot reasonably infer preferences based on the available persona information and should therefore deprioritize such samples. This motivates us to explore verbal uncertainty estimation as a method to filter out cases where persona information is insufficient for the LLM-Judge to make well-informed judgments.

\paragraph{Evaluation of LLMs}
Evaluating natural language generation (NLG) systems is challenging, but the evaluation of LLMs arguably presents even greater difficulties. This is due to the advanced capabilities and versatility of current-generation LLMs, as well as the diverse ways in which they are employed in practice.

Recently, ``LLM-as-Judge" \cite{zheng2023judging} is introduced as a versatile and reference-free evaluation metric that shows high agreement with human annotators on various NLP tasks. Despite concerns over issue such as positional bias, self-enhancement bias, length bias, sensitivity to prompting, and cost \cite{zheng2023judging, stureborg2024large, wu2023style, verga2024replacing, kim2024prometheus}, it is becoming the new paradigm for LLM evaluation \cite{dubois2024alpacafarm,shankar2024validates, liu2024aligning}, and have been used in LLM personalization works such as~\citet{shao_character-llm_2023,andukuri_star-gate_2024}.
However, there is little work in validating LLM-as-a-Personalized-Judge. While MT-Bench~\cite{zheng2023judging} included a role-playing component, it focused only on simplistic cases such as role-playing specific professions and did not account for the complex personas typically used in LLM personalization works, encompassing diverse demographics, user descriptions, and prior interactions. Our work considers more realistic cases of LLM-as-a-Personalized-Judge and carefully examines its validity.

\paragraph{Calibration of LLMs} Pretrained LLMs are well-calibrated but preference tuning can degrade this calibration~\cite{kadavath2022language,achiam2023gpt}. Recent studies have shown that verbalized confidence levels in LLMs are typically more reliable than token-level confidence scores~\cite{tian-etal-2023-just}. Additionally, LLMs possess some intrinsic capabilities to assess the answerability of questions~\cite{kadavath2022language,yin-etal-2023-large}. Building on these findings, our research introduces uncertainty quantification within the context of LLM-as-Personalized-Judge.

\section{Methodology}
In this work, we study LLM-as-a-\textbf{Personalized}-Judge (Figure~\ref{workflow}), building on \citet{zheng2023judging}. We condition an LLM witha persona profile, which can include information ranging from demographic data and socio-behavioral indicators to free-form user descriptions, as well as any other pertinent details that could enrich the persona profile. Using this conditioned persona, we task the LLM with selecting the preferred response to a subjective question in a binary choice setting, aiming to reflect the preferences that the persona would likely have. As is done in~\citet{zheng2023judging}, we also consider a setting where a ``tie'' option is allowed.

As highlighted in Section~\ref{sec:related_work}, 
persona sparsity can lead to instances where an LLM struggles to assess certain questions accurately. However, we hypothesize that the LLM possesses some notion of its uncertainty in these instances. Therefore, we instruct the LLM to estimate the certainty in its answer. The overall workflow is shown in Figure~\ref{workflow}. The prompts used are detailed in Appendix~\ref{prompts}, while the experimental setups are described in Section~\ref{judge_setups}.

\section{Experimental Setup}
\subsection{Datasets}
\textbf{PRISM}~\cite{kirk2024prism} is a participatory, representative, and individualized human feedback dataset. It encompasses feedback on over 8,000 conversations, gathered from 1,500 participants across 75 countries. Additionally, the dataset is enriched with detailed participant profiles.\\
\textbf{OpinionQA}~\cite{santurkar2023whose}, built using Pew Research's American Trends Panel, contains 1498 survey questions spanning 15 topics, the participants' responses, and their demographics.\\
\textbf{Empathetic Conversation (EC)}~\cite{omitaomu2022empathic} consists of 1000 essay responses (both empathy score and textual response) to a news article with their demographics and self-reported personality traits. It further includes dialog interactions between paired participants, enriched with various dialog annotations, such as other-reported empathy levels and turn-by-turn emotion ratings.\\
\textbf{Personal Reddit (PR)}~\cite{staab2023beyond} consists of 500 samples of Reddit posts with their (anonymized) personal attributes, such as location, income, and sex. Unlike other datasets, it is specifically designed to test the ability of LLMs to infer explicit persona attributes. For example, if a post mentions ``I remember watching Twin Peaks after coming home from school'', given that Twin Peaks aired from 1990 to 1991, one could reason that the author of the post is now in the age group of 45-50. Other datasets require annotators to complete tasks and questionnaires, where persona variables may influence responses indirectly but do not explicitly reveal persona information.

\subsection{LLM as Personalized Judge}
\label{judge_setups}
As shown in Figure \ref{workflow}, given a persona, we instruct the LLM to infer the preferred response of the persona.
We have three settings: (1) Standard LLM-as-a-Personalized-Judge: In this setting, the model is directed to make a preference judgment based on the persona, similar to in~\citet{zheng2023judging}. (2) Standard LLM-as-a-Personalized-Judge with Verbal Uncertainty Estimation. In addition to (1), we add an instruction for the model to estimate its certainty in the task on a scale of 1 to 100. (3) Standard LLM-as-a-Personalized-Judge with a Tie Option. In this setting, we introduce a third option, allowing the model to indicate a tie. In this case, we do not permit the model to express verbal uncertainty.

We study the performance of GPT-4~\cite{achiam2023gpt}, GPT-3.5~\cite{openai_introducing_nodate}, Command R+~\cite{cohere_commandrplus}, and LLama3 70B~\cite{llama3}. For generation with all models, we use nucleus sampling~\cite{Holtzman2020The} with top-p of 0.95 and temperature of 0.7. For LLama3 70B, we load the model in 16 bit. In cases when the model reject to answer the question or fail to follow the formatting instruction, We ask the model regenerate at most 4 times until we can parse the results. For details on our experimental setups for each dataset, please refer to Appendix \ref{exp_detail}.

\section{Results}
\begin{table*}[ht!]
\small
\centering
\begin{tabular}{lllcccccccc}
\toprule
\textbf{Setup} & \multicolumn{4}{c}{No Tie (R=50\%)} & \multicolumn{4}{c}{With Tie (R=33\%)} \\
\cmidrule(lr){2-5}\cmidrule(lr){6-9}
\textbf{Model} & Llama3 & GPT-3.5 & GPT-4 & Command R+ & Llama3 & GPT-3.5 & GPT-4 & Command R+ \\ \midrule
PR         &   0.949     &    0.796      &   0.946    &   0.964     &   -   &    -     &    -    &      -   \\ \midrule
PRISM           &   0.722    &     0.656    &    0.728   &    0.720     &   0.678    &     0.537    &   0.727    &    0.689     \\ \midrule
OpinionQA    &    0.629      &     0.569    &   0.635    &     0.616    &    -   &     -    &    -   &     -    \\ \midrule
EC           &    0.507   &     0.529    &    0.591   &    0.541     &   0.376      &   0.384      &   0.417    &   0.430  \\ \midrule \midrule
Average & 0.702 & 0.638 & 0.725 & 0.710 & 0.527  &  0.461 &   0.572  & 0.560

\\ \bottomrule
\end{tabular}
\caption{Agreement between different LLM judges with the human ground truth from PRISM, OpinionQA, and EC. Following \citet{zheng2023judging}, we report two cases for the judge: with tie and without tie.  The agreement between two random judges under each setup is denoted as ``R=''. Average is calculated as the direct (non-weighted) average of accuracy across the datasets. Due to the unavailability of relevant data in the PR and OpinionQA datasets, we thus omit them for the with tie setting.}
\label{acc}
\end{table*}

\begin{table*}[ht]
\centering
\small
\begin{tabularx}{\textwidth}{@{}c*{8}{Y}@{}}
\toprule
\textbf{Model} & \multicolumn{2}{c}{Llama 3} & \multicolumn{2}{c}{GPT-3.5} & \multicolumn{2}{c}{GPT-4} & \multicolumn{2}{c}{Command R+} \\
\cmidrule(lr){2-3} \cmidrule(lr){4-5} \cmidrule(lr){6-7} \cmidrule(lr){8-9}
\textbf{Confidence} & High & Low & High & Low & High & Low & High & Low \\ 
\midrule \midrule
PR & \cellcolor[HTML]{C0C0C0} 0.948 (492/520) & \textit{1.000 \ \ \  (5/5)} & \cellcolor[HTML]{C0C0C0} 0.792 (375/473) & 0.667 (34/51) & \cellcolor[HTML]{C0C0C0} 0.942 (228/243) & 0.950 (266/280) & \cellcolor[HTML]{C0C0C0} 0.958 (345/361) & 0.976 (160/164) \\ 
\midrule
PRISM & \cellcolor[HTML]{C0C0C0} 0.753 (570/758) & 0.625 (150/240) & \cellcolor[HTML]{C0C0C0} 0.673 (520/773) & 0.599 (135/227) & \cellcolor[HTML]{C0C0C0} 0.908 (108/120) & 0.703 (612/871) & \cellcolor[HTML]{C0C0C0} 0.893 (133/149) & 0.690 (587/852) \\ 
\midrule
OpinionQA & \cellcolor[HTML]{C0C0C0} 0.706 (964/1366) & 0.566 (928/1640) & \cellcolor[HTML]{C0C0C0} 0.568 (1641/2890) & 0.578 (58/102) & \cellcolor[HTML]{C0C0C0} 0.804 (385/480) & 0.602 (1526/2535) & \cellcolor[HTML]{C0C0C0} \textit{1.000 \ \ (2/2)} & 0.616 (1856/3013) \\ 
\midrule
EC & \cellcolor[HTML]{C0C0C0} 0.504 (240/478) & 0.548 (16/31) & \cellcolor[HTML]{C0C0C0} 0.530 (250/472) & 0.517 (14/29) & \cellcolor[HTML]{C0C0C0} \textit{1.000  \   \ (4/4)} & 0.588 (295/502) & \cellcolor[HTML]{C0C0C0} \textit{-  \ \ \ \ \ \ \ \ \ (0/0)} & 0.541 (276/510) \\ 
\bottomrule
\end{tabularx}
\caption{Agreement for high and low confidence for different models. ``High'' and ``low'' refers to the certainty level estimated by the model. The number of correct answers/total number of samples are provided below the accuracy. In our analysis, we use a certainty threshold of 80 to classify responses as high confidence. The italicized numbers indicate that very few samples are available for accuracy calculation.}
\label{high_low_conf}
\end{table*}

\paragraph{LLM-based Personalized Judge shows low agreement with human}
In Table \ref{acc}, we present the agreement between different LLM judges and the human ground truth. The results indicate that, for binary preference choice questions where random guessing would yield an accuracy of 50\%, the average accuracy of the LLM-as-a-Personalized-Judge, even for the most powerful model, is only 72.5\%. This accuracy is significantly lower than the 80+\% agreement reported in \citet{zheng2023judging}, and it drops to around 60\% for challenging tasks such as EC and OpinionQA. These findings suggest that LLM judges are less reliable for personalization tasks compared to simpler role-playing tasks.

Accuracy also varies substantially across different tasks and LLMs. PR is the easiest task, with all models performing best on this dataset; for instance, GPT-4 achieves an accuracy of 94.6\%.
This high performance is likely attributable to  the dataset's design, where one response explicitly reflects certain persona characteristics while the other does not. Thus, PR may not represent genuine personalization. For example, if a persona includes a statement like ``I enjoy outdoor activities,'' and one response is ``I love hiking,'' while the other is "I prefer watching movies indoors," the distinction is clear. Hence, PR may not reflect personalization; rather, it can be reviewed as a task akin to instruction-following and textual entailment. 

Conversely, EC appears to be the most difficult, with all models achieving less than 60\% accuracy. This may be because the persona included lacks sufficient predictive power for the task. The articles in EC are chosen to elicit empathetic responses, which are generally very negative and lead to similar responses from different individuals.

Among different models, GPT-4 consistently performs the best across nearly all tasks, followed by Command R+ and Llama-3 70B. In contrast, GPT-3.5 shows substantially worse performance.

\begin{figure*}[htbp]
\centering
\subfloat[]{
        \includegraphics[width=0.33\textwidth]{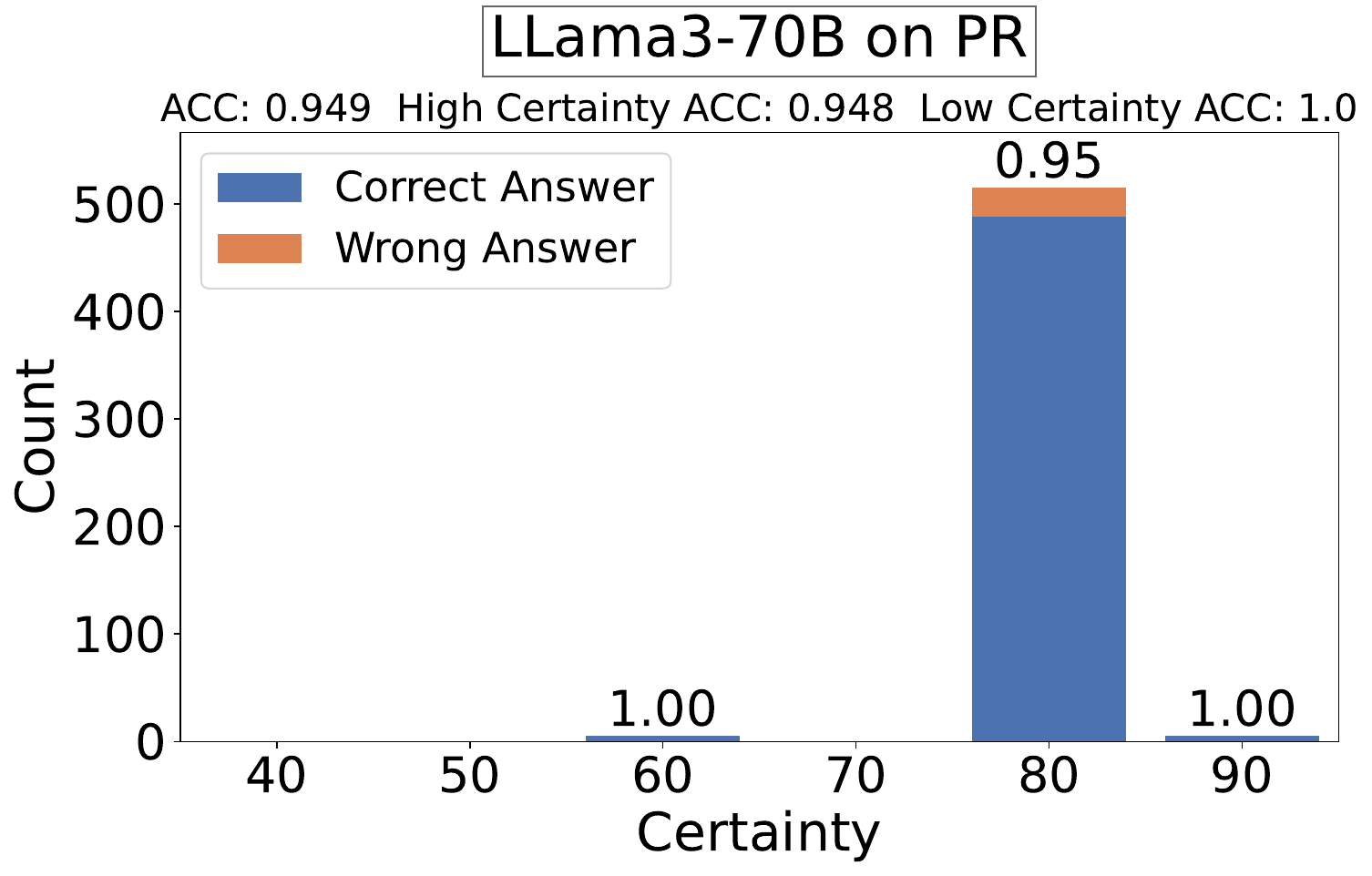}
    }
    \subfloat[]{
        \includegraphics[width=0.33\textwidth]{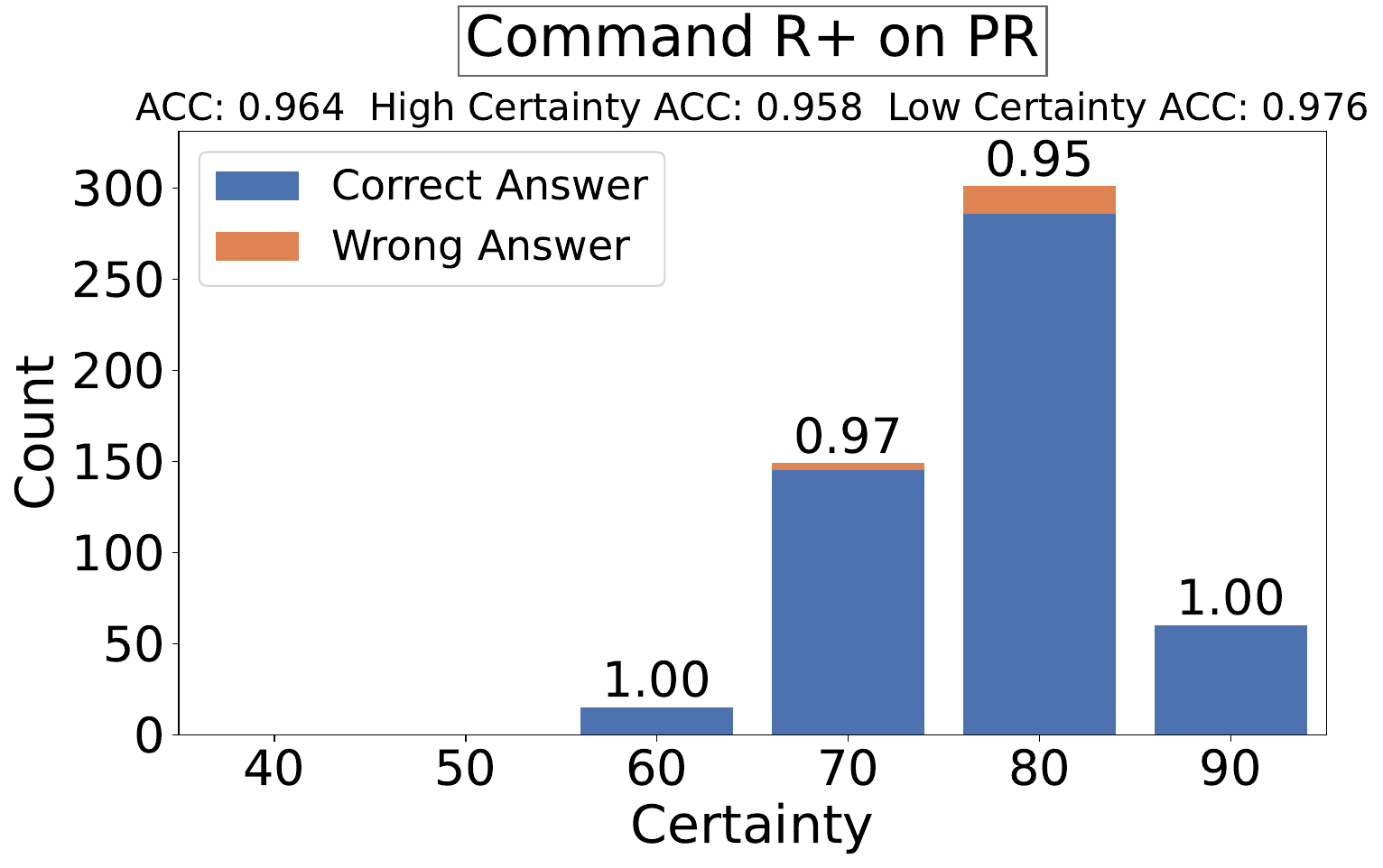}
    }
    \subfloat[]{
        \includegraphics[width=0.33\textwidth]{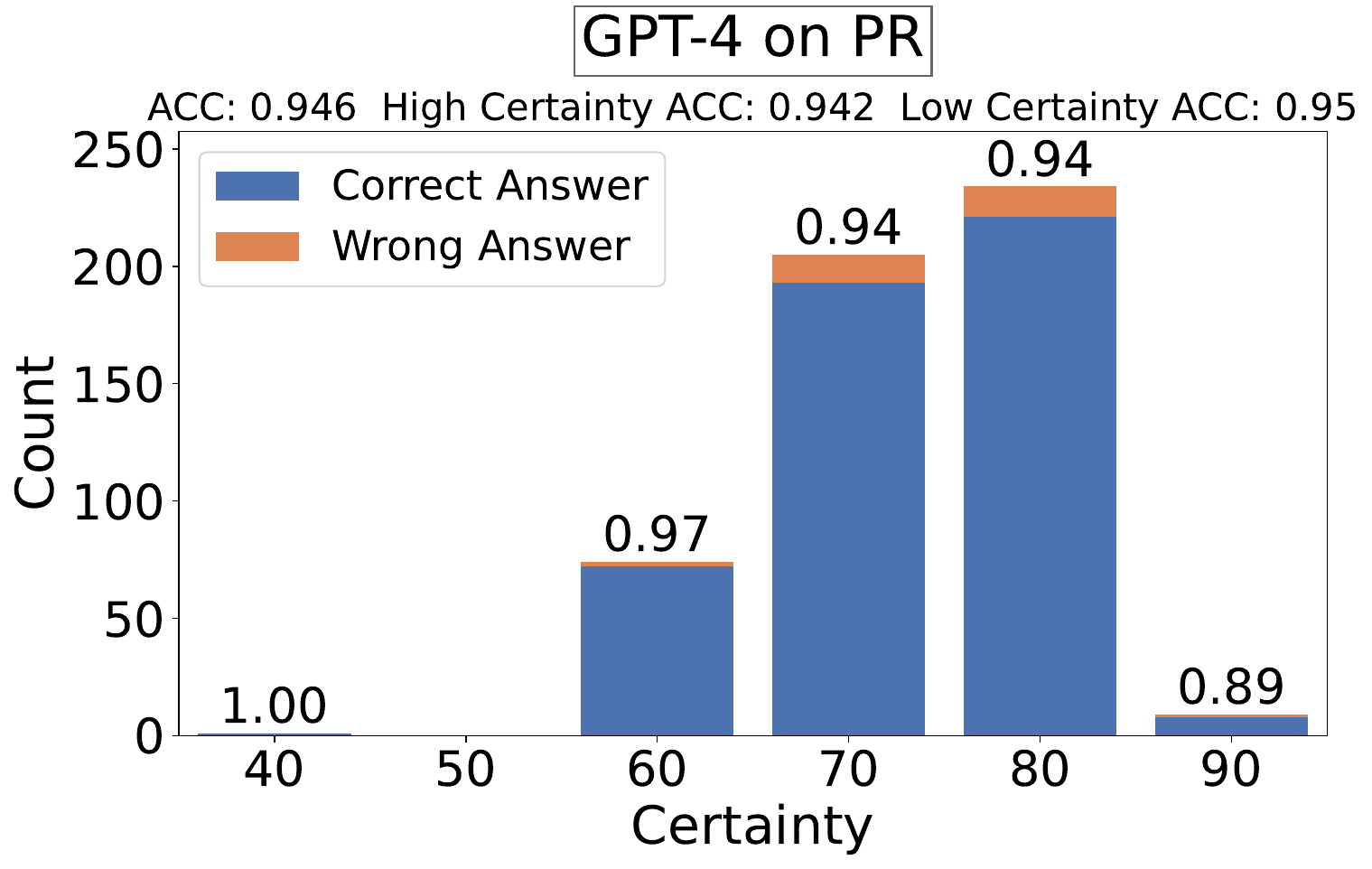}
    }    \\
\centering
    \subfloat[]{
        \includegraphics[width=0.33\textwidth]{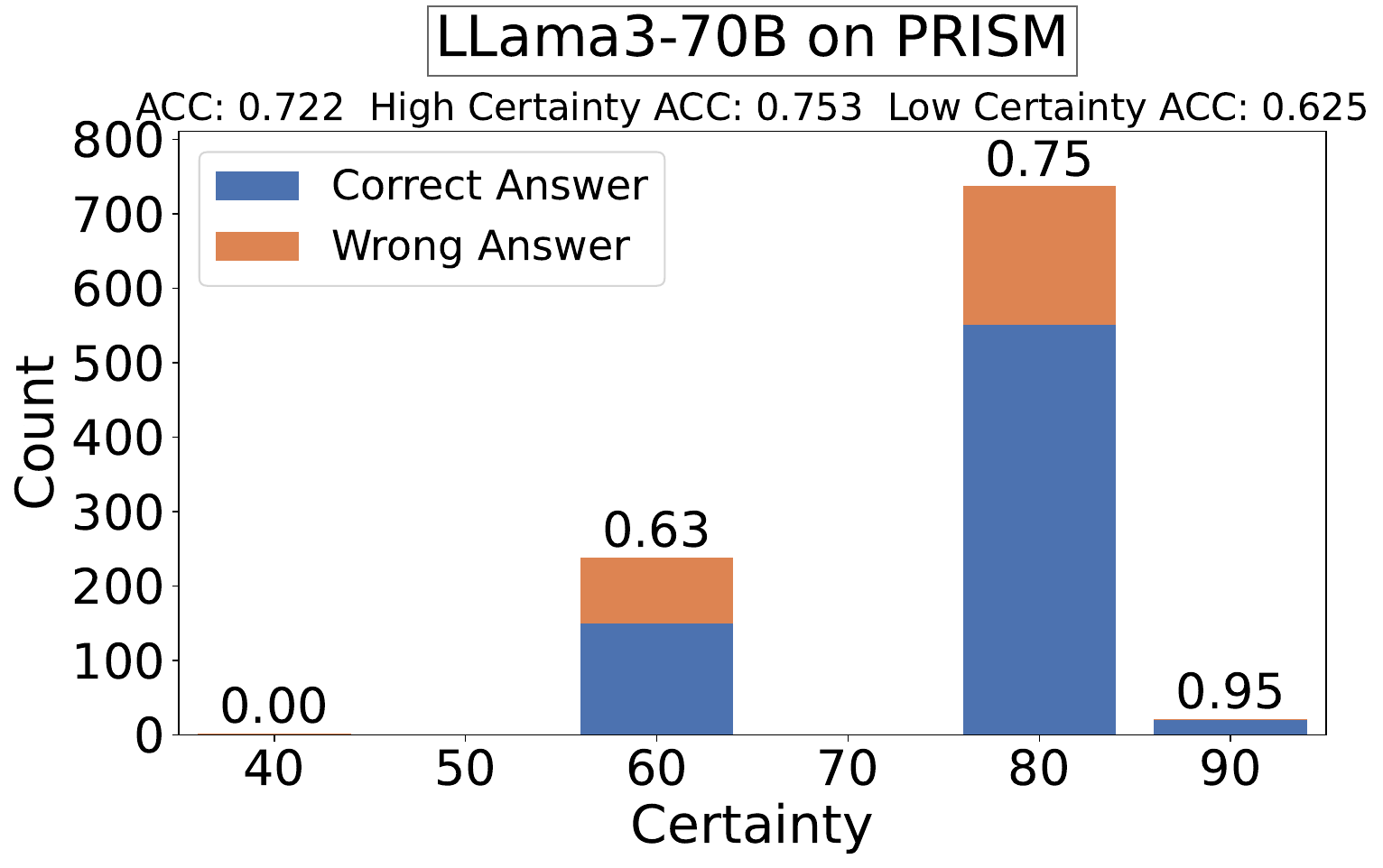}
    }
    \subfloat[]{
        \includegraphics[width=0.33\textwidth]{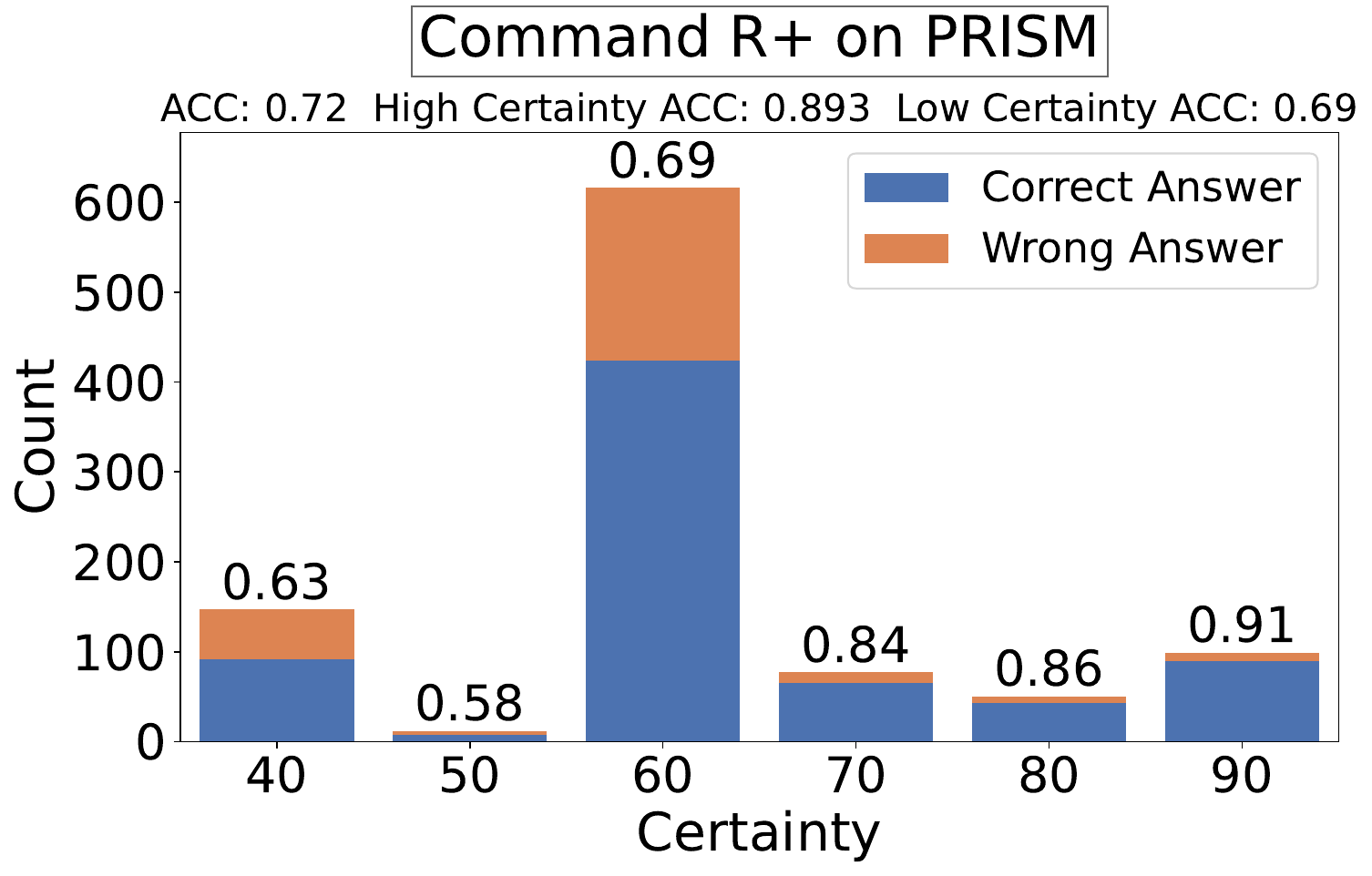}
    }
    \subfloat[]{
        \includegraphics[width=0.33\textwidth]{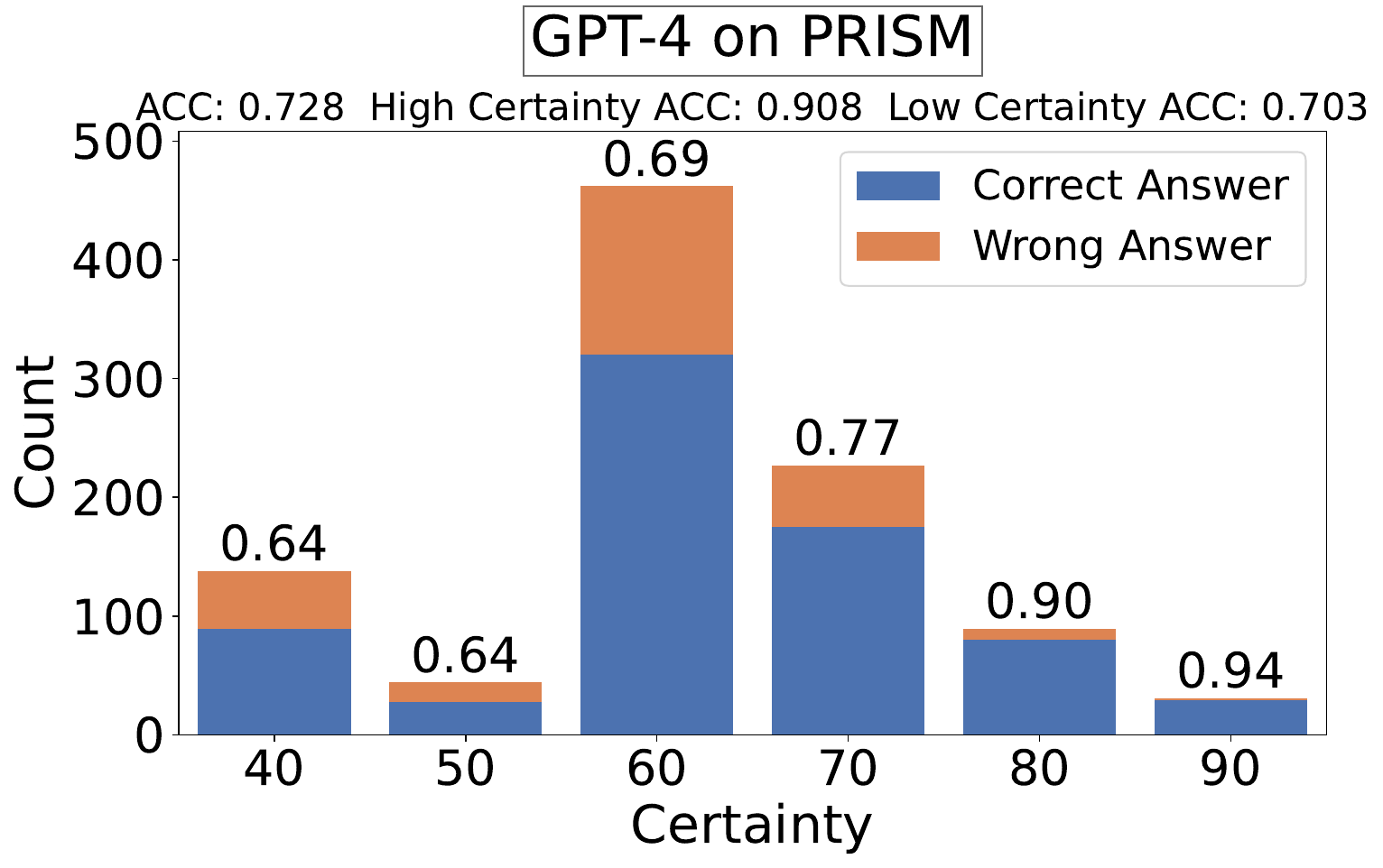}
    }\\
\centering
    \subfloat[]{
        \includegraphics[width=0.33\textwidth]{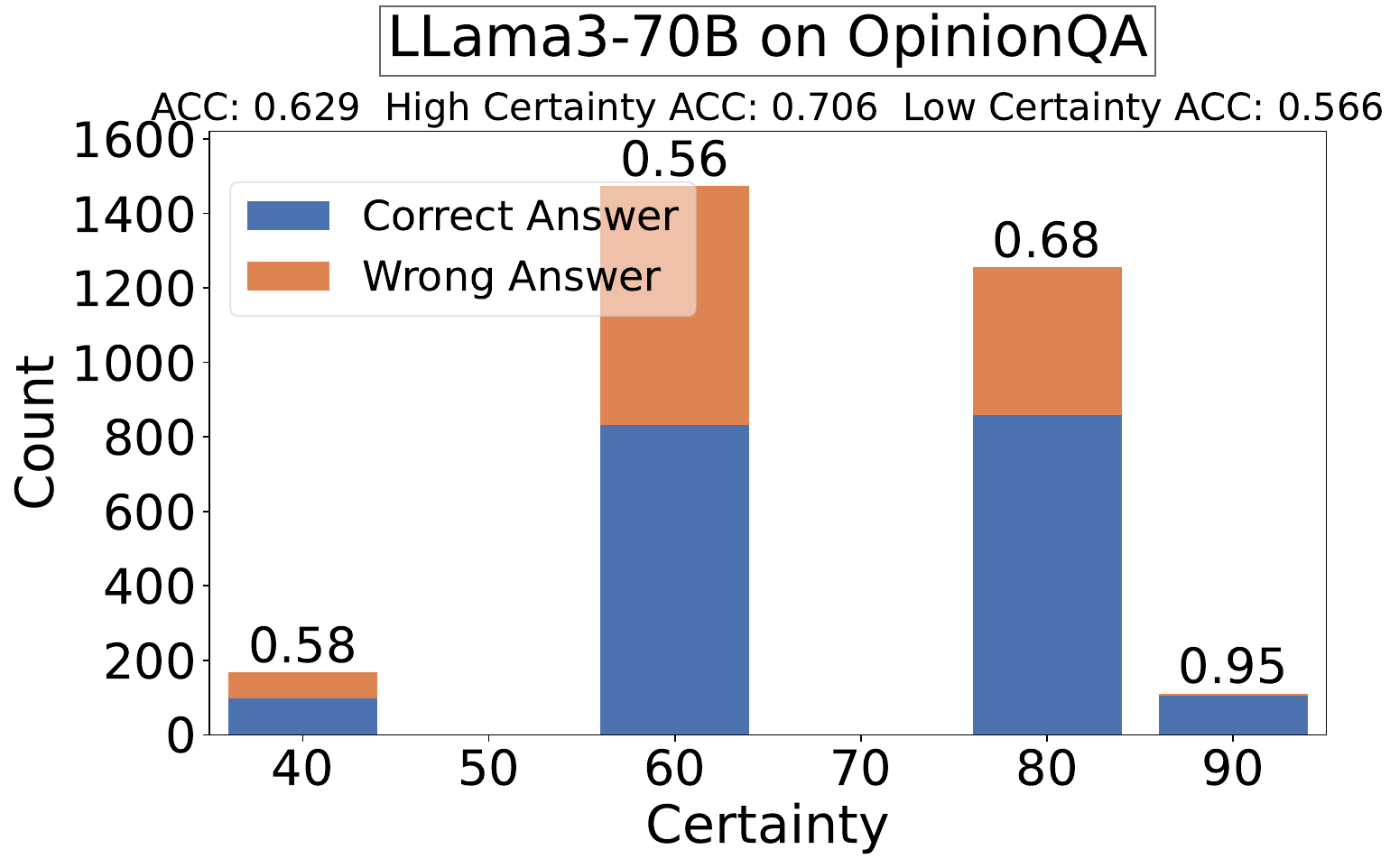}
    }
    \subfloat[]{
        \includegraphics[width=0.33\textwidth]{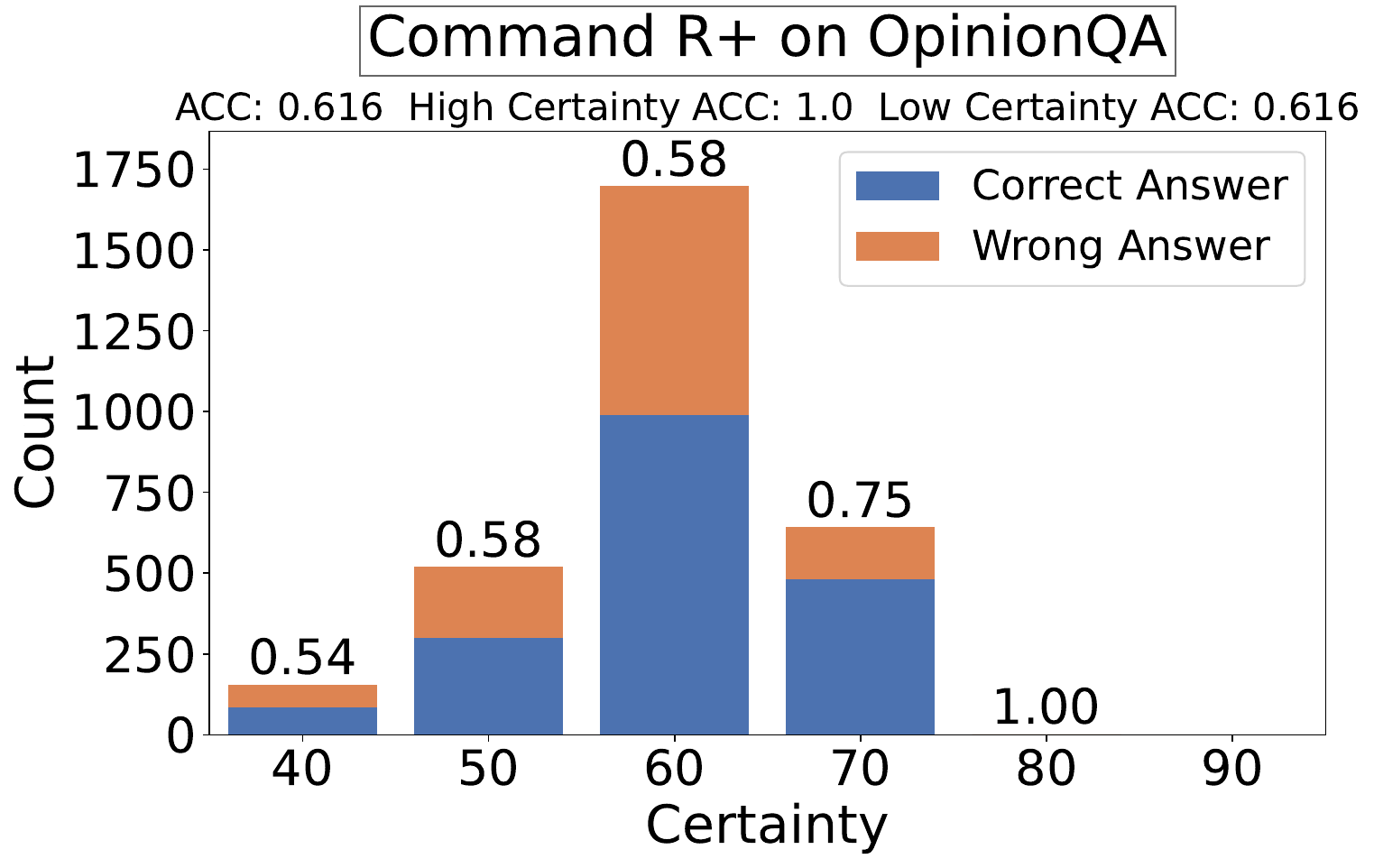}
    }
    \subfloat[]{
        \includegraphics[width=0.33\textwidth]{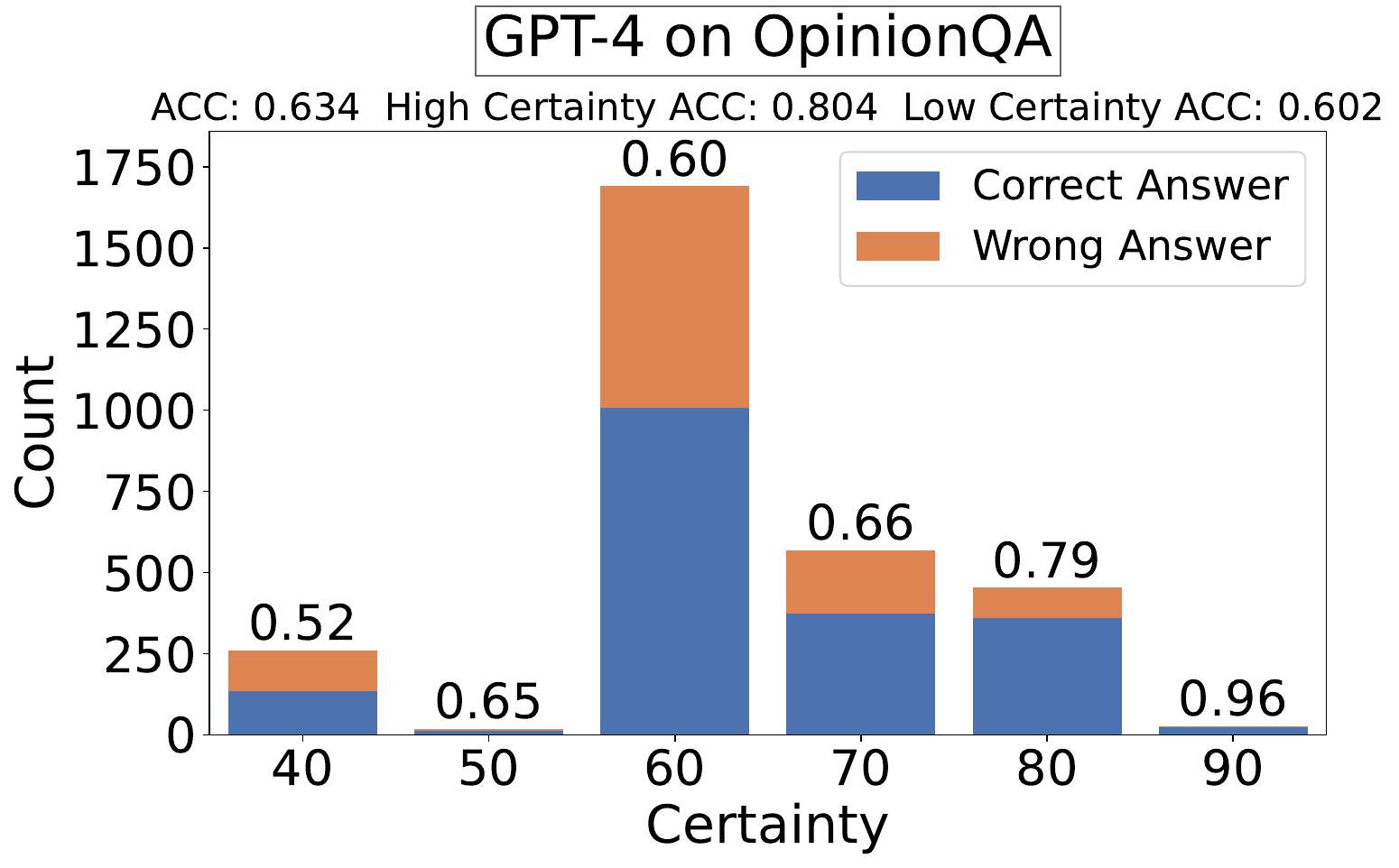}
    } \\
\centering
\subfloat[]{
        \includegraphics[width=0.33\textwidth]{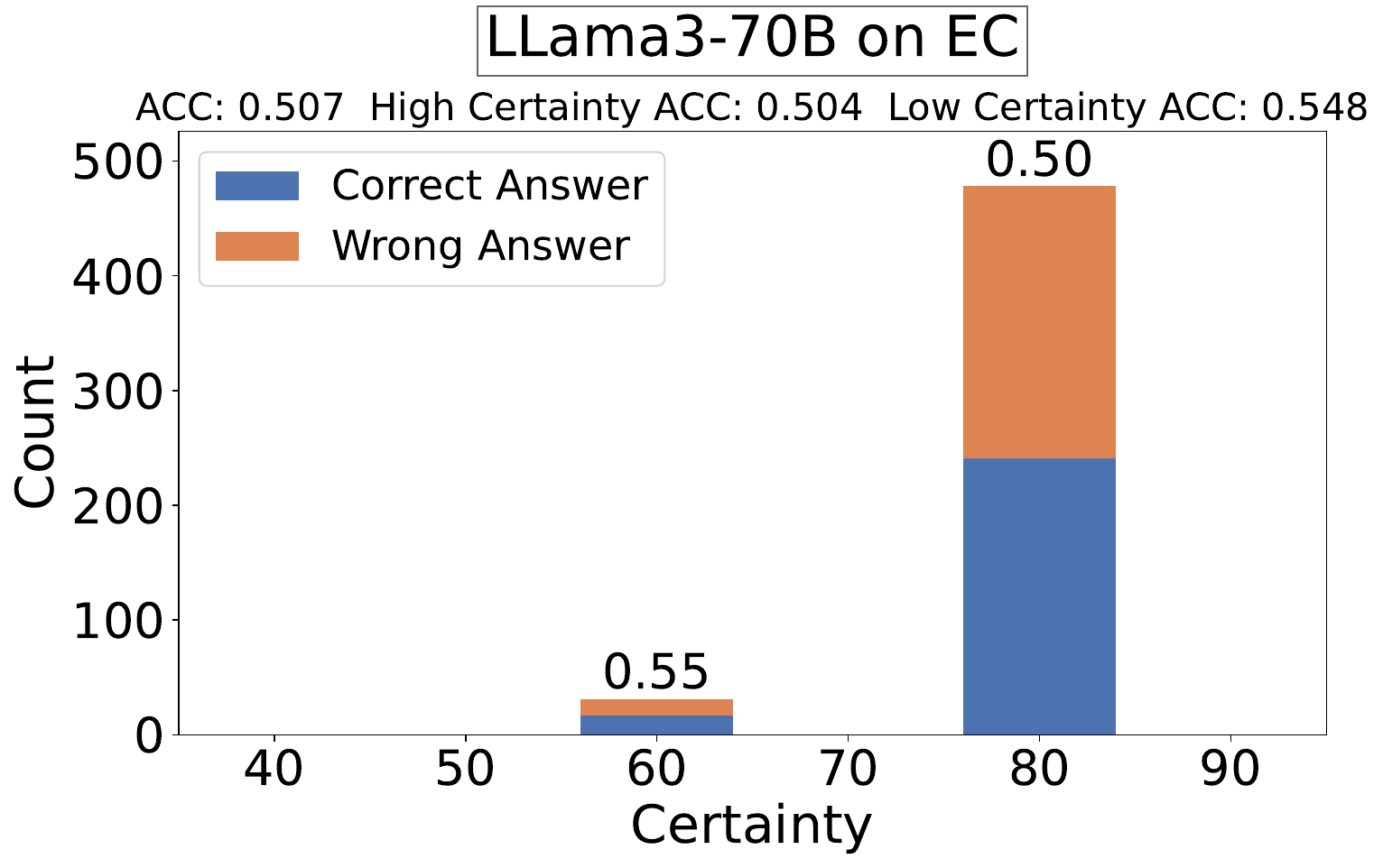}
    }
    \subfloat[]{
        \includegraphics[width=0.33\textwidth]{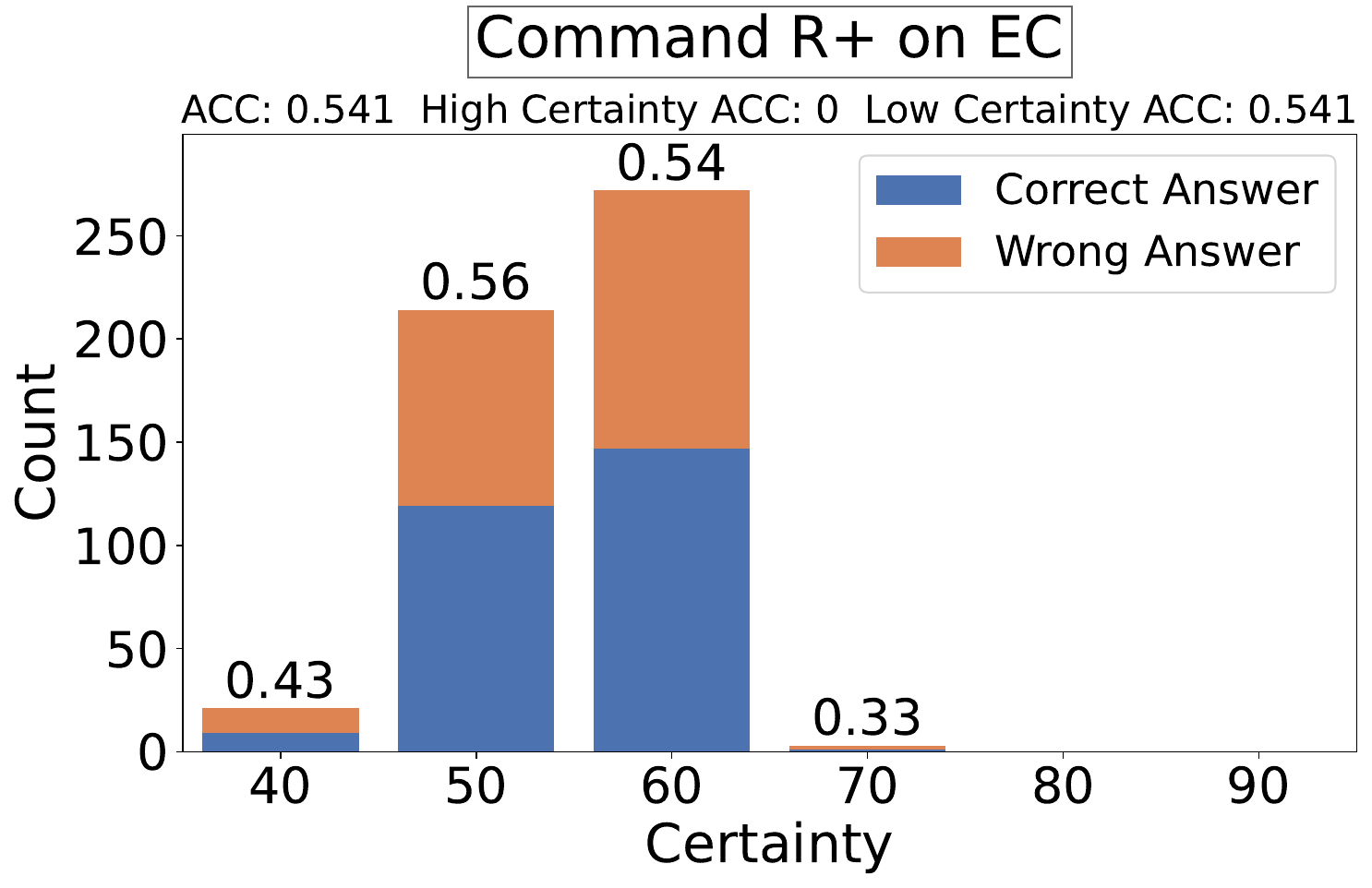}
    }
    \subfloat[]{
        \includegraphics[width=0.33\textwidth]{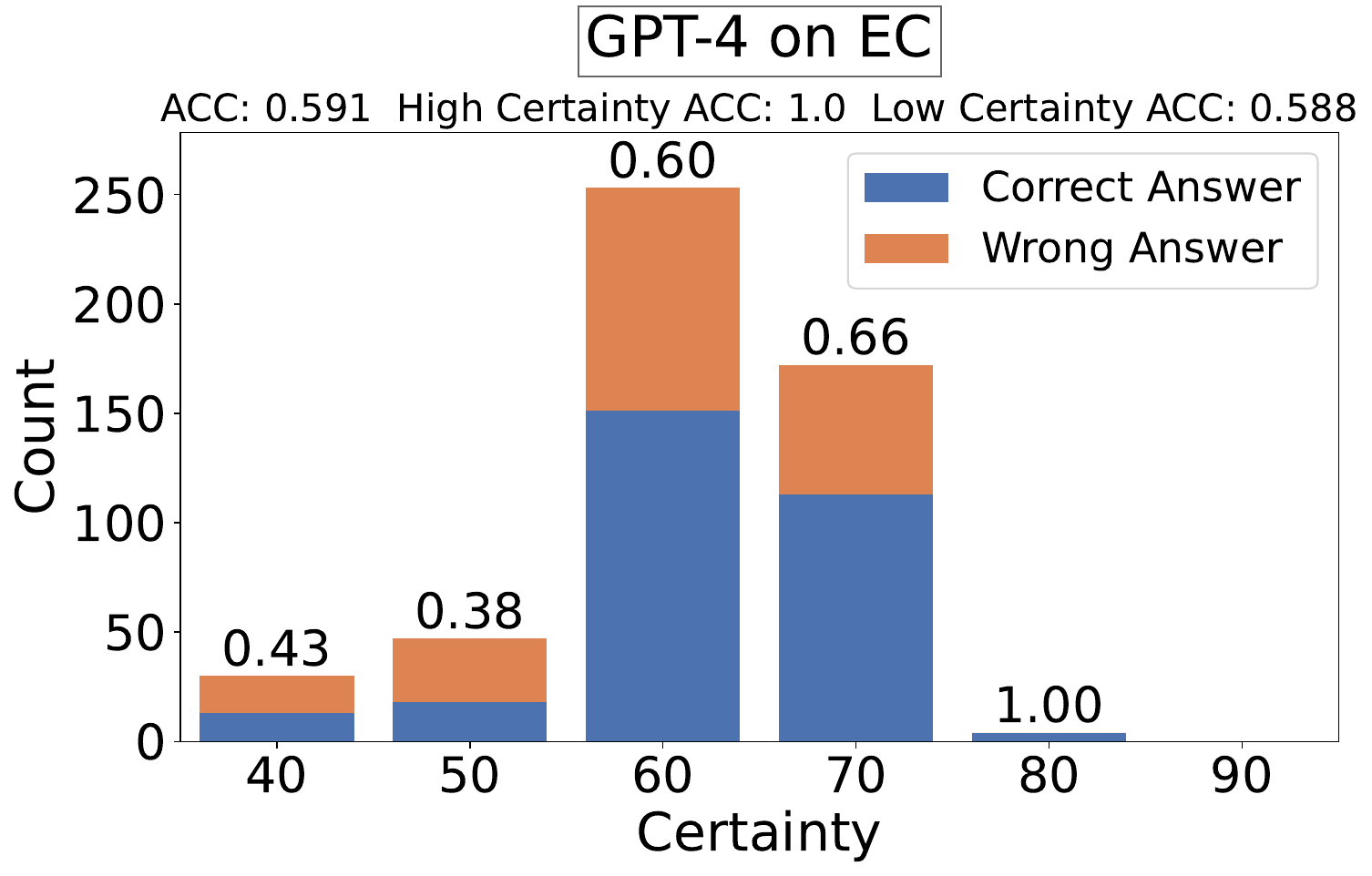}
    }
\caption{\textbf{Distribution of LLM verbal certainty score and the corresponding accuracy.} The plots show the certainty distribution and corresponding accuracy of correct (blue) and wrong (orange) answers for LLAMA3-70B, Command R+, and GPT-4 models on PR, PRISM, OpinionQA, and EC datasets. Each plot provides overall accuracy (ACC), high certainty accuracy (High Certainty ACC), and low certainty accuracy (Low Certainty ACC). The top of each bar shows the accuracy within that certainty bin. The certainty levels are truncated to be between 40 and 90 by adjusting values outside this range.}
\label{certainty_dist}
\end{figure*}

When models are allowed to choose a tie option, similar trends are observed. While model performance on both PRISM and EC declines, the drop is much more significant for EC. This is because models rarely choose the tie option even when it is available. Therefore, we suggest that incorporating a tie option in practical applications is not ideal. Conceptually, using tie options to filter samples is not as flexible as having the model express its confidence since we can choose different thresholds to control the number of samples being filtered. Additionally, for PR and OpinionQA, we do not add a tie option because we do not have ground truth data for ties. 
\paragraph{Certainty estimation improves Personalized Judge}
In Figure \ref{certainty_dist}, we plot the accuracy of predictions across different certainty levels for various models and tasks. Some models, such as Llama3, exhibit a highly concentrated distribution of certainty levels within a narrow range, while others display a more Gaussian-like distribution, which is arguably more ideal. We observe a clear trend indicating that predictions from more powerful LLMs (e.g. GPT-4) with higher certainty scores are more likely to be correct. In contrast, less powerful LLMs, (e.g.GPT-3.5), often struggle to accurately quantify their uncertainty. This observation suggests that we can rely, at least to some degree, on a model's self-assessed confidence to evaluate whether the information in the persona is sufficient for making reliable predictions.

We manually assign a threshold of 80 for all models to classify a sample as high-confidence and we show in Table \ref{high_low_conf} the judge performance for each model under this certainty thresholding.  High confidence samples from GPT-4 and Command R+ can achieve approximately 80\% agreement with human ground-truth, on par with~\citet{zheng2023judging}.\\\\
\paragraph{LLM-as-a-Personalized-Judge performance varies greatly across models}
We also observe substantial performance differences across models. As shown in Table \ref{acc}, GPT-4 is the most powerful model followed by Command R+ and then LLama-3 70B. The performance of GPT-3.5 is significantly worse, with a 5\%--10\% performance gap on average. More importantly, GPT-3.5 and LLama-3 70B's capacity to self-estimate certainty is significantly worse. As shown in Table \ref{high_low_conf} and Figure \ref{certainty_dist}, GPT-3.5 fails to achieve higher accuracy on high-confidence samples. LLama-3 70B has slightly better certainty estimation than GPT-3.5 but is still far from GPT-4 and Command R+ which achieve 80\%+ accuracy on high-confidence samples. Given these results, we focus the rest of our experiment and discussion primarily on GPT-4 and Command R+.

\begin{table}[t!]
\centering
\tiny
\begin{tabularx}{0.95\columnwidth}{@{}c*{4}{Y}@{}}
\toprule
\textbf{Model} & \multicolumn{2}{c}{GPT-4} & \multicolumn{2}{c}{Command R+} \\
\cmidrule(lr){2-3} \cmidrule(lr){4-5}
\textbf{Confidence} & High & Low & High & Low \\ 
\midrule \midrule
All Features & \cellcolor[HTML]{C0C0C0} 0.804 (385/480) & 0.602 (1526/2535) & \cellcolor[HTML]{C0C0C0} \textit{1.000 \ \ (2/2)} & 0.616 (1856/3013) \\
\midrule
Three Imp. Features & \cellcolor[HTML]{C0C0C0} 0.833 (199/239) & 0.593 (1646/2776) & \cellcolor[HTML]{C0C0C0} \textit{1.000 \ \ \ (2/2)} & 0.612 (1843/3013) \\ 
\midrule
Least Imp. Feature & \cellcolor[HTML]{C0C0C0} \textit{0.400 \ (4/10)} & 0.589 (1769/3005) & \cellcolor[HTML]{C0C0C0} \textit{0.000 \ \  \ (0/1)} & 0.546 (1645/3014)  \\ 
\bottomrule
\end{tabularx}
\caption{Ablation study on using different features of the user persona to predict the user preference on OpinionQA. The italicized numbers indicate that too few
samples are used to compute the accuracy.}
\label{OpinionQA}
\end{table}

\begin{table}[t!]
\centering
\tiny
\begin{tabularx}{0.95\columnwidth}{@{}c*{4}{Y}@{}}
\toprule
\textbf{Model} & \multicolumn{2}{c}{GPT-4} & \multicolumn{2}{c}{Command R+} \\
\cmidrule(lr){2-3} \cmidrule(lr){4-5}
\textbf{Confidence} & High & Low & High & Low \\ 
\midrule \midrule
All Features & \cellcolor[HTML]{C0C0C0} 0.908 (108/120) & 0.703 (612/871) & \cellcolor[HTML]{C0C0C0} 0.893 (133/149) & 0.690 (160/164) \\
\midrule
Three Imp. Features & \cellcolor[HTML]{C0C0C0} 0.904 (104/115) & 0.680 (594/873) & \cellcolor[HTML]{C0C0C0} 0.737 (225/305) & 0.653 (454/696) \\ 
\midrule
Least Imp. Feature & \cellcolor[HTML]{C0C0C0} 0.880 (81/92) & 0.71 (642/903) & \cellcolor[HTML]{C0C0C0} 0.780 (166/214) & 0.644 (506/787) \\ 
\bottomrule
\end{tabularx}
\caption{Ablation study on using different features of the user persona to predict the user preference on PRISM.}
\label{PRISM}
\end{table}

\begin{table}[t!]
\centering
\tiny
\begin{tabularx}{0.95\columnwidth}{@{}c*{4}{Y}@{}}
\toprule
\textbf{Method} & \multicolumn{2}{c}{GPT-4} & \multicolumn{2}{c}{Third Person Human Judge} \\
\cmidrule(lr){2-3} \cmidrule(lr){4-5}
\textbf{Confidence} & High & Low & High & Low \\ 
\midrule \midrule
All Features & \cellcolor[HTML]{C0C0C0} 0.792 (38/48) & 0.592 (149/252) & \cellcolor[HTML]{C0C0C0} 0.714 (30/42) & 0.620 (160/258) \\
\midrule
\textbf{Overall Average} & \multicolumn{2}{c}{0.623 (187/300)} & \multicolumn{2}{c}{0.633 (190/300)} \\
\midrule
\bottomrule
\end{tabularx}
\caption{Third-person human evaluation on OpinionQA: Crowd annotators assess the preferences of individuals based on specific profile descriptions, and these assessments are compared with the GPT-4 powered LLM-as-a-Personalized-Judge. For each sample, three annotators provide annotations, and the final human answer is determined by a simple majority vote.}
\label{human_eval}
\end{table}
\paragraph{Confidence distribution as a proxy of task and sample difficulty}
In Table \ref{high_low_conf}, we observe significant variation in the number of samples categorized under high and low confidence across different tasks. We hypothesize that this variation corresponds to the difficulty of the tasks. For example, as shown in Table \ref{acc}, PR is the most straightforward task based on high average accuracy for most models while EC poses significant challenges for all models. Thus, as shown in Table \ref{high_low_conf} and Figure \ref{certainty_dist}, on the PR dataset, around 50\% of the prediction by GPT-4 and nearly 100\% predictions by Command R+ is considered high confidence, much higher than the PRISM and OpinionQA datasets, which has only around 10\% - 20\% high confidence samples. On the contrary, only around 1\% of the predictions on EC are considered high-confidence. This result illustrates that on more difficult tasks, LLMs are able to assign low confidence for a larger number of predictions, 
supporting our hypothesis that the an LLM's confidence judgment can be a reliable indicator of task difficulty and persona sparsity. We believe this is a crucial property to have for an LLM-Judge: in personalization tasks, end users may not always be aware of the difficulty level of a given task for all samples. They can therefore rely on the model's confidence as a surrogate measure. When evaluating a personalization task using an LLM-as-a-Personalized-Judge, users should prioritize high-confidence samples, as these are more likely to reflect accurate and reliable judgments. Implementing a confidence threshold can facilitate more meaningful comparisons between methods of personalization in future evaluations.

\paragraph{Certainty significantly drops when only very few persona features are given} In real-world applications, the availability of persona variables can vary, and it is important to observe how the model's confidence changes with both the quantity and relevance of these variables. To explore this, we conduct an ablation study to further verify that LLMs would indeed assign low confidence to the predictions when the persona is insufficiently predictive. We provide different numbers of persona variables to the LLM-Judge.  While the precise predictive power of a persona is hard to quantify, fewer features should lead to lower confidence in LLM predictions. Concretely, instead of using all features as before, we provide the LLM with only three important features (education, location, ethnicity) or one less important feature (religion) for OpinionQA and PRISM.

For OpinionQA, in Table \ref{OpinionQA}, we find that GPT-4 assigns low confidence to much fewer samples when only three or one features are provided. Specifically, the number of high-confidence samples drop  from 480 (16.0\%) to 10 (0.3\%) for the one-feature case. For Command R+, since the number of high-confidence samples is already very low, it remains relatively unchanged. Figure \ref{certainty_different_features} provides a more detailed analysis of the change in certainty distribution when providing a different number of persona variables.

For PRISM, as shown in Table \ref{PRISM}, we observe a similar trend, albeit with fewer changes compared to OpinionQA. We attribute this to the significant inherent quality differences between the two response options provided for each question in the dataset. In many instances, these differences are so pronounced that the preferred choice remains evident regardless of the persona (as indicated in \ref{no_persona}). Consequently, the number of high-confidence samples predicted by GPT-4 only slightly decreased (from 120 to 92) from full-feature to one-feature, while maintaining a high accuracy on these high-confidence samples.
\paragraph{LLM-as-a-Personalized Judge achieves comparable performance as third-person human judge}

In dialog system personalization, third-person human annotation is a widely adopted evaluation strategy. Typically, this involves human crowd annotators inferring the prefrences of personas of others rather than expressing their own opinions. Although this is considered a gold standard, its effectiveness scenarios remain underexplored.

For our evaluation, we use the OpinionQA dataset and collect crowd annotations via Prolific. We sample 300 instances, with each instance receiving annotations from three different human annotators, totaling 30 samples per annotator. Annotators infer how a persona would respond in specific scenarios and rate their certainty levels, using the same prompts as the LLMs. We establish the final human answer based on a simple majority vote, and average the certainty levels of the majority answers to establish ground truth certainty. The crowd-sourced results are presented in Table~\ref{human_eval}.
The overall accuracy of GPT-4 was 62.3\%, closely matching the human-level accuracy of 63.3\%. On high-certainty samples, GPT-4 achieved an accuracy of 79.2\%, surpassing the human performance level of 71.4\%. These results corroborate findings by~\citet{rescala2024can} which suggests that LLMs can match human performance in evaluating whether arguments are likely to resonate with individuals characterized by specific persona attributes.

To further validate the reliability of the crowd judgments, we conducted bootstrap sampling 1,000 times with 30 samples each, performing random draws without replacement of the annotations. The mean agreement between two annotators is 0.597, with a standard deviation of 0.087, indicating a reasonably high level of internal consistency in our results.
Additionally, we provide the unaggregated annotation results in Table~\ref{human_eval_unaggregate}. Here, human performance was inferior to the majority vote, likely reflecting variations in annotators' skills.
Our human annotation results also underscore the inherent challenges in personalization evaluation. While first-person annotations can be considered ground truth and are therefore always accurate, even third-person human judges often struggle to reach correct judgments in many cases.

Our crowd-sourcing exercise indicates that LLMs when used as personalized judges, can achieve accuracy levels comparable to those of human annotators. However, under the default setting, the overall accuracy remains low, likely due to persona sparsity issues. When certainty thresholding is applied, LLMs achieve better accuracy on high-certainty samples than human annotators. While we advocate for the collection of more first-person datasets—where individuals provide information about themselves and then answer questions—we also propose that LLMs, with certainty thresholding, represent a promising and scalable alternative for evaluating personalization tasks in the absence of first-person data.

\section{Conclusion}
In this paper, we formalized and examined the validity of LLM-as-a-Personalized-Judge. Contrary to previous assumptions, we demonstrated that the standard LLM-as-a-Judge setting is not sufficiently reliable for personalization tasks, showing low agreement with human ground truth.  We identified persona sparsity as a major cause of this unreliability. We then introduced verbal certainty estimation and found that powerful LLMs (e.g. GPT-4) are capable of effectively assessing the certainty of their own responses. This led to the observation that high-certainty samples indeed exhibit high accuracy (80\%). We additionally conducted a human annotation experiment and found that LLM-as-a-Personalized-Judge achieves comparable accuracy as third-person human judge and surpasses humans on high-certainty samples. 
While we advocate for the collection of more first-person personalization data, we also believe that a certainty-aware LLM-as-a-Personalized-Judge is a promising proxy for evaluation, particularly in cases first-person preference data are not available, provided that personas are as fine-grained as possible. We hope our work helps the community recognize the challenges in evaluating LLM personalization and ultimately leads to the development of LLMs that better serve each individual's preferences and needs.

\section*{Limitations}

The availability of diverse and comprehensive datasets for evaluating LLM personalization remains limited, and such datasets are predominantly available in English. Consequently, we cannot make conclusive statements about the performance of LLMs as personalized judges in non-English languages. Furthermore, existing multilingual LLMs often exhibit cultural gaps~\cite{liu2023multilingual}, which suggests that their performance might be suboptimal in non-English contexts due to the complex cultural associations tied to persona variables. Future research should aim to compile and utilize more extensive datasets with richer and more varied persona attributes in a multilingual setting to better evaluate and improve LLM personalization.

Although numerous methods for quantifying uncertainty in LLMs have been proposed, we opted to use direct verbal estimation. This method is straightforward and has better performance compared to the model's conditional probability~\cite{tian-etal-2023-just}. Although a comprehensive evaluation of existing uncertainty estimation in LLM-as-a-Personalized-Judge would make a valuable contribution for future work, it is beyond the scope of the present work, which is mostly focused on the integration of uncertainty estimation into LLMs-as-Personalized-Judge as a framework.

\section*{Ethical Considerations}
The goal of the LLM-as-a-Personalized-Judge is to enhance personalization in LLMs to better serve a diverse global community. However, achieving this goal necessitates a rigorous adherence to ethical principles throughout the research and production phases. For example, personalization should always remain an opt-in choice for end users, ensuring user autonomy and consent without any adverse consequences for those who opt out. Additionally, LLMs have been shown to have various kinds of social biases~\cite[\textit{inter alia}]{pmlr-v139-liang21a,hu2023generative,LIU2022103654}, some of which may exhibit itself during the LLM-as-a-Judge process. We need to be mindful of such biases not to reinforce the bias and stereotypes encoded in the LLMs. Privacy concerns become especially salient when personal information is utilized to fine-tune or condition models. It is crucial to manage such data responsibly by obtaining explicit user consent and adhering to data protection regulations, such as the General Data Protection Regulation. In our research, we have relied on existing publicly available datasets, which have undergone institutional review board approval and anonymization prior to release.

\section*{Acknowledgements}
T.H is supported by Gates Cambridge Trust (grant OPP1144 from the Bill \& Melinda Gates Foundation). This work was partially performed using resources provided by the Cambridge Service for Data Driven Discovery (CSD3) operated by the University of Cambridge Research Computing Service (www.csd3.cam.ac.uk), provided by Dell EMC and Intel using Tier-2 funding from the Engineering and Physical Sciences Research Council (capital grant EP/T022159/1), and DiRAC funding from the Science and Technology Facilities Council (www.dirac.ac.uk). We are grateful for support received in the form of research access or credits from Cohere and OpenAI. We thank Yinhong Liu, Ivan Vulić, Songbo Hu, and Fabian David Schmidt for helpful feedback and discussions at various stages of the project.

\bibliography{anthology,custom}

\appendix
\section{Appendix}

\begin{table*}[ht]
\small
\centering
\begin{tabular}{ll}
\toprule
Task      & Persona Variables                                                                                     \\
\midrule \midrule
PR        & Age, Sex, Living Country, Birth Country, Education, Occupation, Income, Marital Status                 \\ \midrule
PRISM     & Age, Sex, Race, Birth Country, Living Country, Employment Status, Education, Marital Status, Religion  \\ \midrule
OpinionQA & Age, Sex, Living Country, Education, Citizenship, Marital Status, Religion, Party, Ideology, Income \\ \midrule
EC        & Age, Sex, Race, Education, Income, Big Five Personality Traits     \\
\bottomrule
\end{tabular}
\caption{Persona variables used for different tasks}
\label{persona_vars}
\end{table*}

\begin{table}[t!]
\centering
\tiny
\begin{tabularx}{0.95\columnwidth}{@{}c*{4}{Y}@{}}
\toprule
\textbf{Method} & \multicolumn{2}{c}{GPT-4} & \multicolumn{2}{c}{Third Person Human Judge} \\
\cmidrule(lr){2-3} \cmidrule(lr){4-5}
\textbf{Confidence} & High & Low & High & Low \\ 
\midrule \midrule
All Features & 0.792 (114/144) & \cellcolor[HTML]{C0C0C0} 0.592 (447/755) & 0.644 (94/146) & \cellcolor[HTML]{C0C0C0} 0.587 (442/753) \\
\midrule
\textbf{Overall Average} & \multicolumn{2}{c}{0.624 (561/899)} & \multicolumn{2}{c}{0.596 (536/899)} \\
\midrule
\bottomrule
\end{tabularx}
\caption{Third-person human evaluation on OpinionQA: Crowd annotators assess the preferences of individuals based on a specific profile descriptions, and these assessments are compared with the GPT-4 powered LLM-as-a-Personalized-Judge.
}
\label{human_eval_unaggregate}
\end{table}

\subsection{LLM Model Details}
For GPT-4o, we use gpt-4o-2024-05-13. For GPT-3.5-turbo, we use gpt-3.5-turbo-0125.

\subsection{Experiment Details}
\label{exp_detail}
For experiments on PRISM, we run the experiments on the first 1,000 samples from the utterance subset of the dataset. We only consider the first turn in each conversation. As suggested by \citet{kirk2024prism}, we consider the two responses with scores smaller or equal to 10 to be a tie. For setting (1) and (2), we only consider samples that is not deemed a tie. For (3), we include those tie samples as well. To mitigate positional bias, we randomly shuffle the position of the two responses. To mitigate self-enhancement bias (preferring text generated by itself)~\cite{zheng2023judging}, we filter out the responses that are generated by the same LLM as the Judge. We also filter out the responses that refuse to answer the question. This is because different LLMs have different safety constraints and different rejection ratios but humans typically find the LLM rejection undesirable and assign low scores to it.

For experiments on OpinionQA, we randomly select one binary choice question from each of the 15 topics covered by OpinionQA. For each question, we randomly select 200 respondent's answers.

For experiments on EC, we only consider the essay response part of the dataset. We select two responses to a news article, and let the LLM to infer which is written by a user with a specific persona. We ran 500 samples in total. We consider two essay responses to be a tie if the difference in their empathy score or distress score is smaller than 2. Since most responses are similar in score and are considered as tie, we control the ratio of the tie cases in EC to be the same as the ratio in PRISM which is around 20\% in setting 3). 

For experiments on PR, since no user responds to the same question, we need to provide a question-response pair and let the LLM to infer which response is likely written by the target user. Concretely, for each persona-question-response triple, we select the most similar persona to the target user based on cosine similarity computed by the all-MiniLM-L6-v2\footnote{https://huggingface.co/sentence-transformers/all-MiniLM-L6-v2} model from Sentence Transformer \cite{reimers-gurevych-2019-sentence}, the backbone of which is a MiniLM model~\cite{NEURIPS2020_3f5ee243}.  Then take this user's response to another question to form another persona-question-response triple and let the LLM infer which response is written by the target user.

\subsection{Persona Variables Used for Each Task}
In Table \ref{persona_vars}, we show the persona variables we used for each task.

\subsection{Crowdsourcing Details}
\label{crowdsourcing_detail}
We recruited 30 U.S. annotators via Prolific. For quality control purposes, each annotator was required to have completed a minimum of 50 prior crowd tasks with an approval rating of at least 99\%. We applied the quota sample feature from Prolific to ensure that the gender and political affiliation distribution among annotators was balanced. We restricted the age of the participants to be between 18 and 75 years old. Annotators were compensated at the rate of \$13.5 per hour. This study received approval from an institutional ethics review board.

\subsection{Results With and Without Persona}
\begin{table}[H]
\small
\centering
\begin{tabular}{l c c}

\toprule
          & w/ Persona & w/o Persona \\ \midrule \midrule
PR     &    -       &      -     \\ \midrule
PRISM     &    0.728        &     0.685       \\ \midrule
OpinionQA &     0.635       &        0.575     \\ \midrule
EC        &      0.591      &        0.498     \\ \bottomrule
\end{tabular}
\caption{Accuracy when predicting the user preference with and without user persona on our three subjective tasks. Experiments are done with Command R+. PR is omitted because it is infeasible to conduct an experiment without a persona on the PR dataset.}
\label{no_persona}
\end{table}

\subsection{Number of Persona Variables Provided Influence Certainty Distribution}
\begin{figure*}[htbp]
\centering
\subfloat[]{
        \includegraphics[width=0.33\textwidth]{figure/prism_all_features_gpt-4o_certainty_plot.pdf}
    }
    \subfloat[]{
        \includegraphics[width=0.33\textwidth]{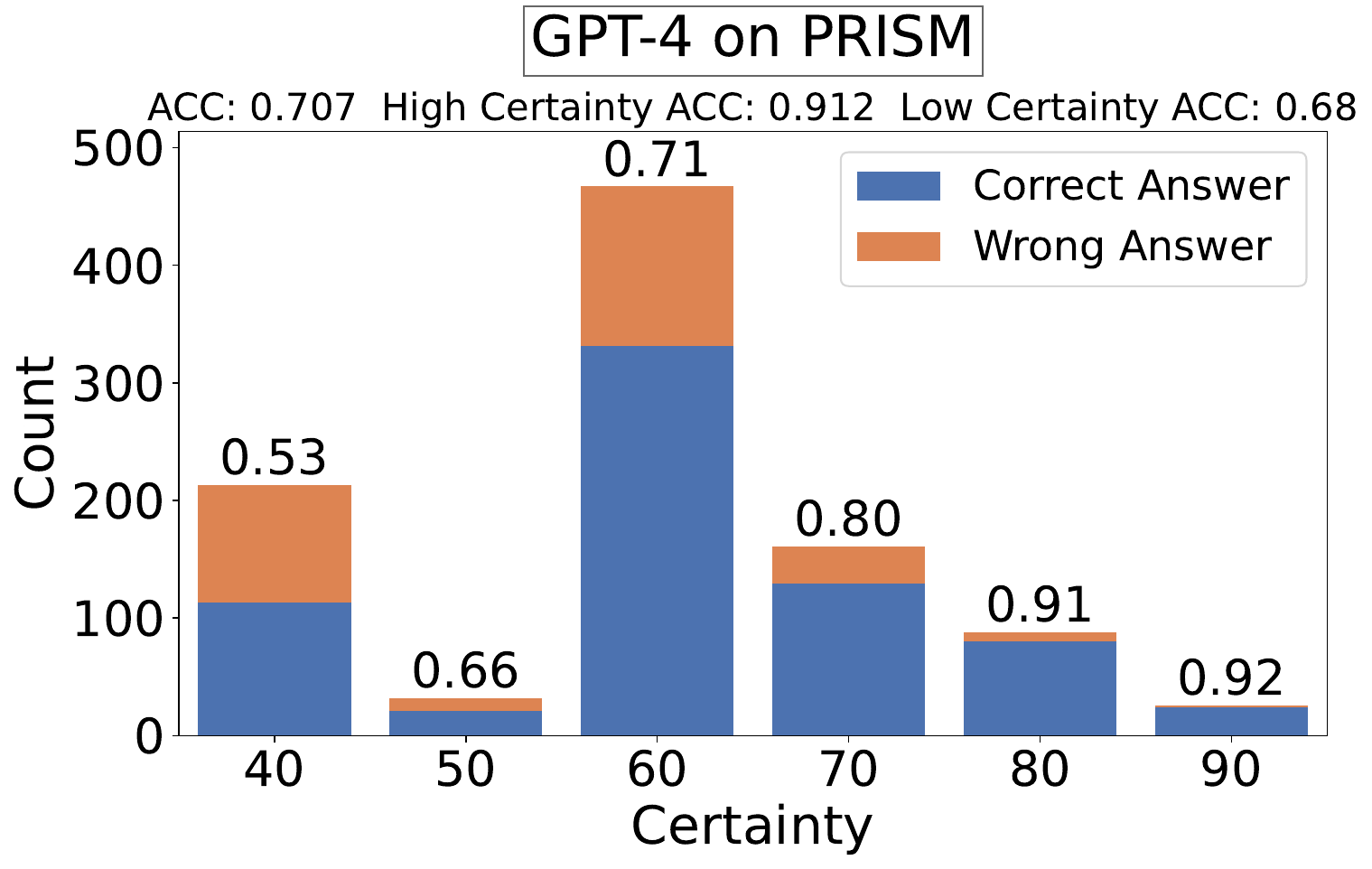}
    }
    \subfloat[]{
        \includegraphics[width=0.33\textwidth]{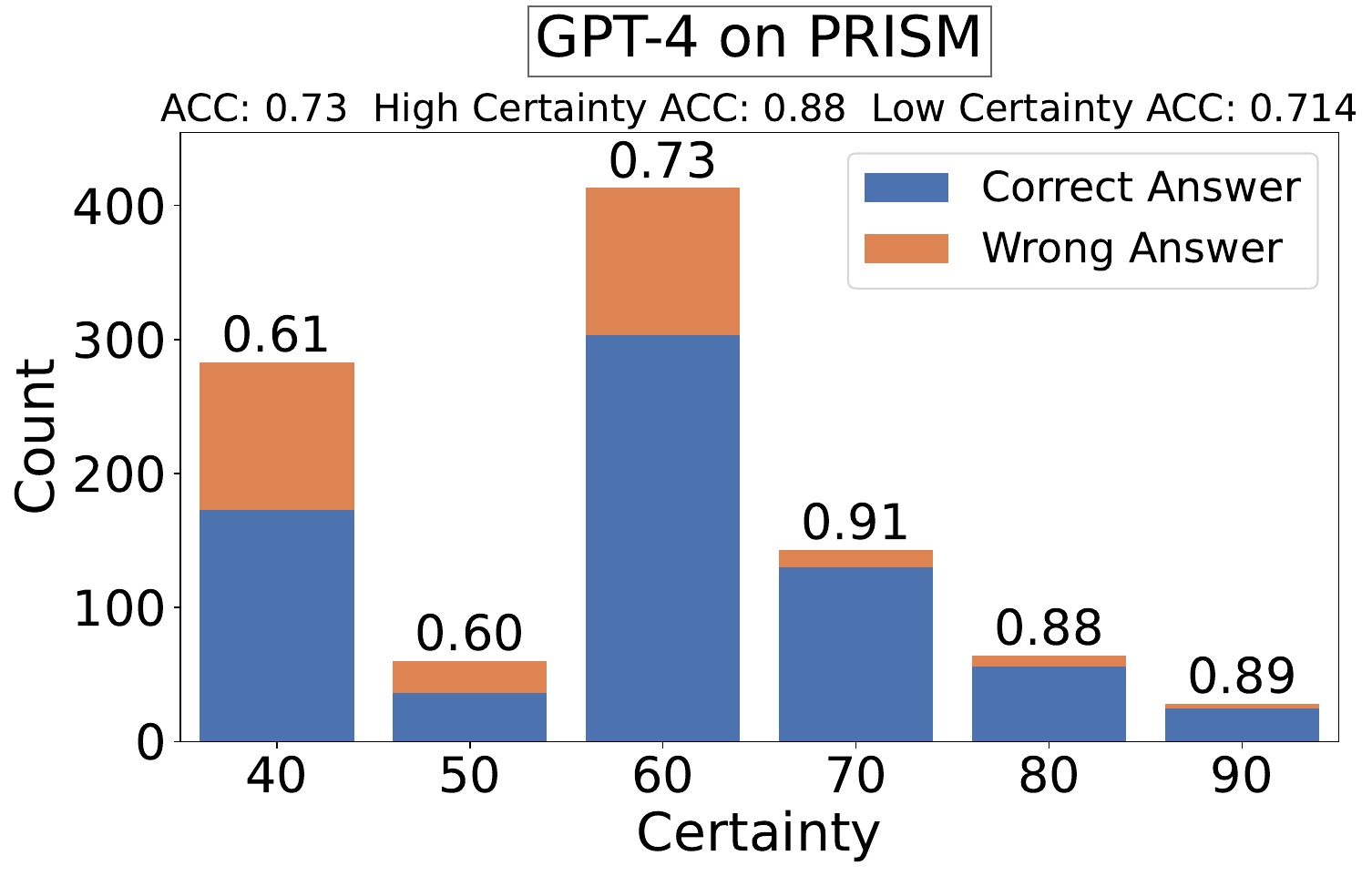}
    } \\
\centering
\subfloat[]{
        \includegraphics[width=0.33\textwidth]{figure/prism_all_features_command-r-plus_certainty_plot.pdf}
    }
    \subfloat[]{
        \includegraphics[width=0.33\textwidth]{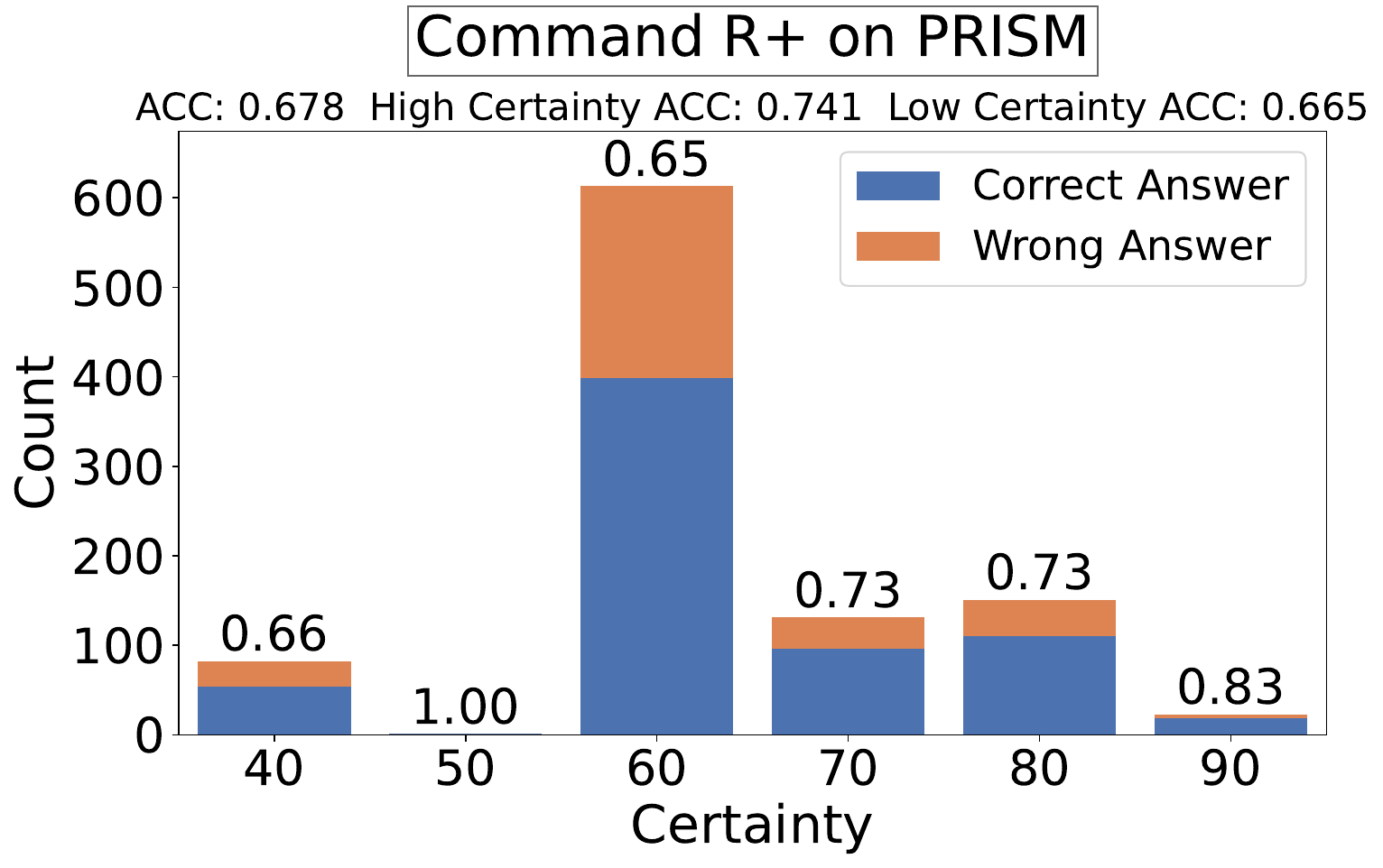}
    }
    \subfloat[]{
        \includegraphics[width=0.33\textwidth]{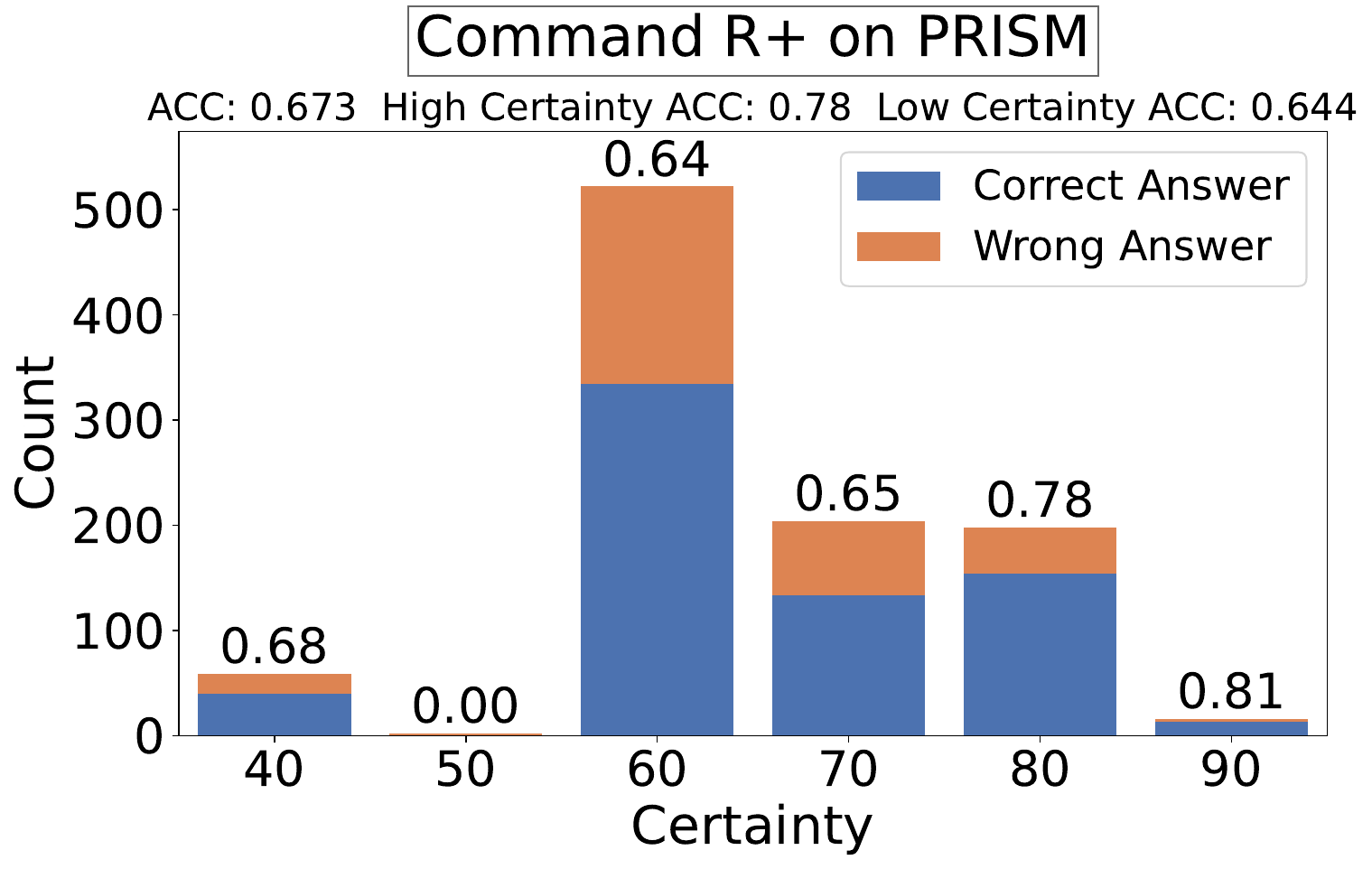}
    } \\
\centering
\subfloat[]{
        \includegraphics[width=0.33\textwidth]{figure/opinionqa_all_features_all_features_gpt-4o_jun_11_certainty_plot.pdf}
    }
    \subfloat[]{
        \includegraphics[width=0.33\textwidth]{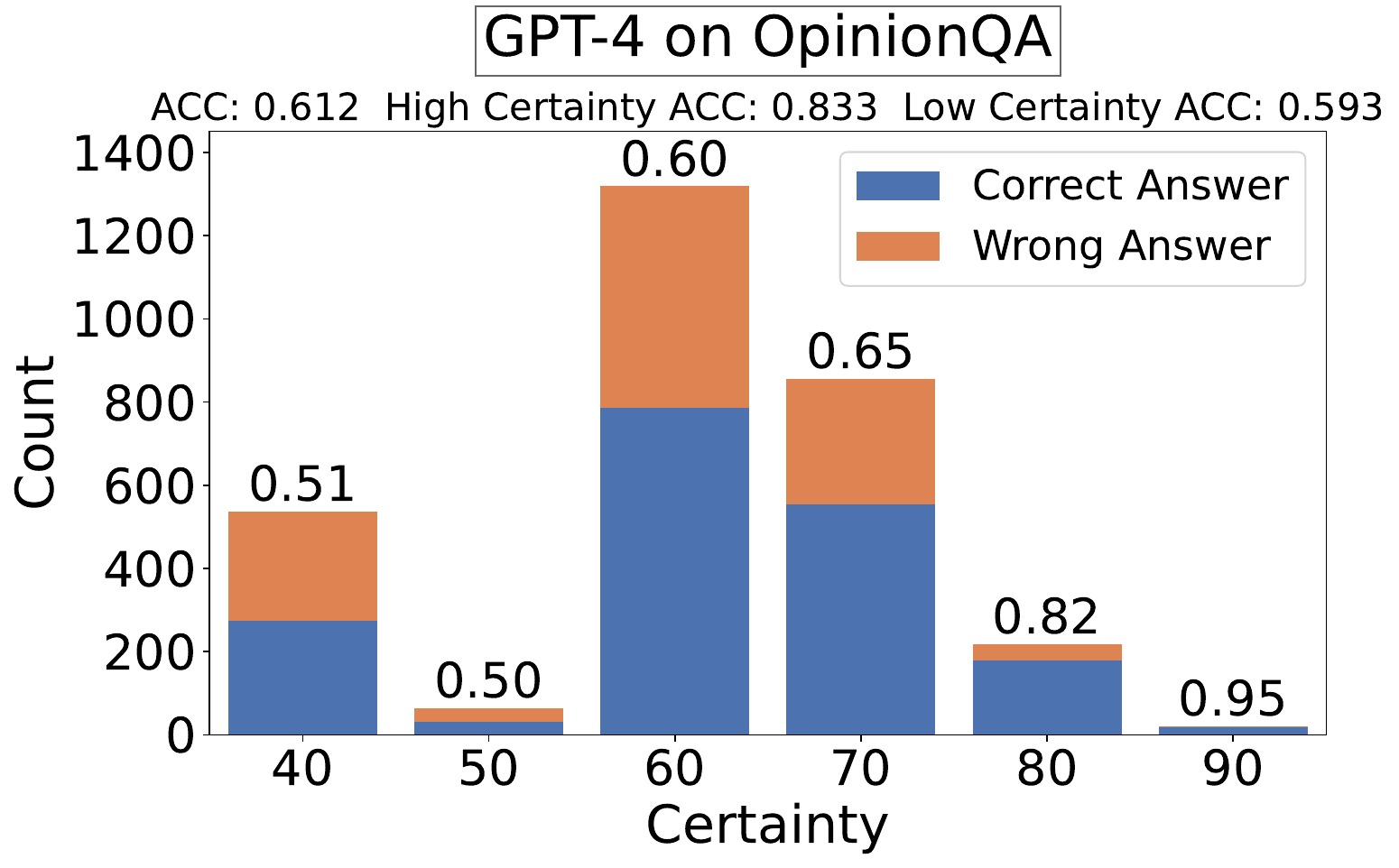}
    }
    \subfloat[]{
        \includegraphics[width=0.33\textwidth]{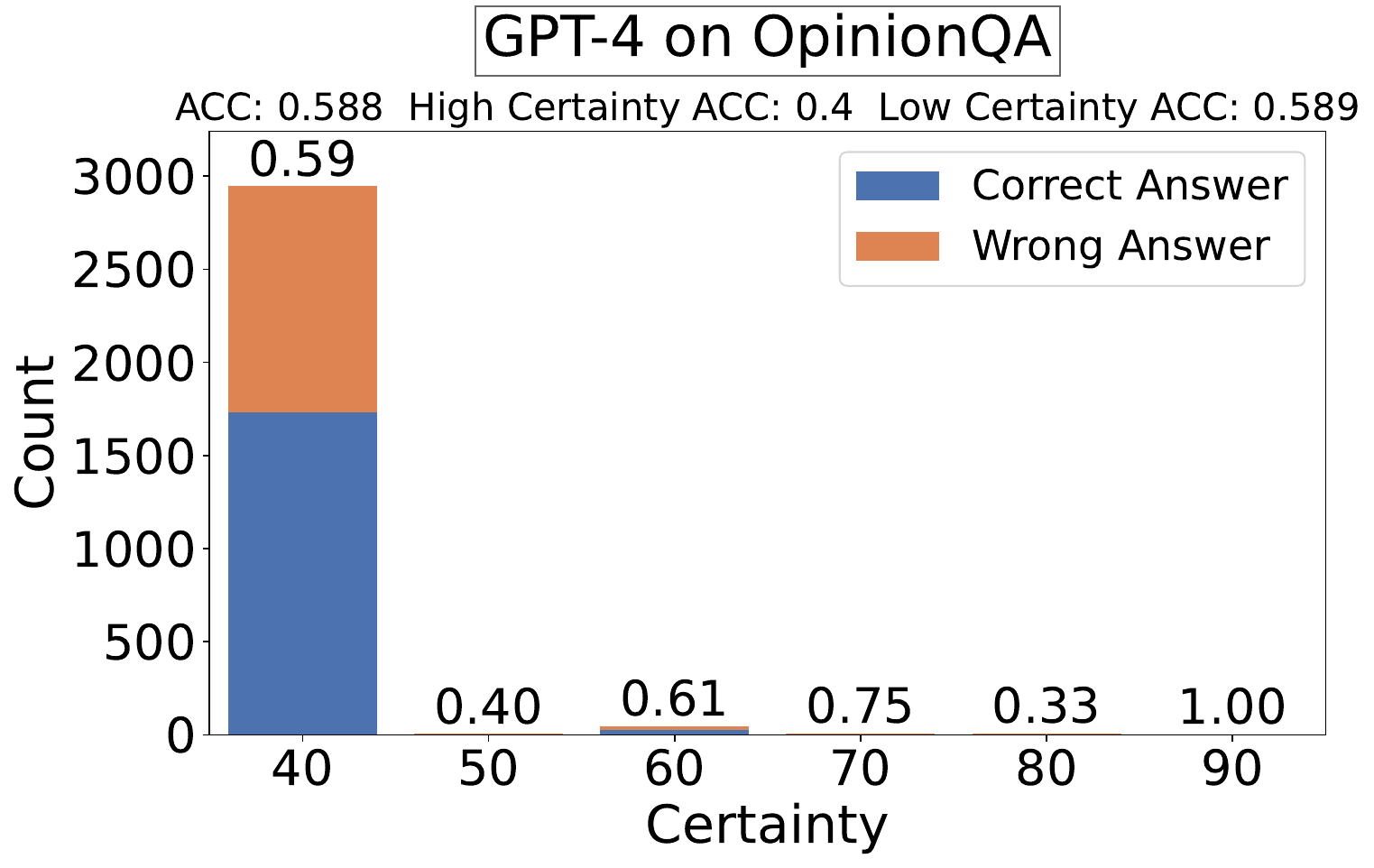}
    } \\
\centering
\subfloat[]{
        \includegraphics[width=0.33\textwidth]{figure/opinionqa_all_features_all_features_command-r-plus_jun_11_certainty_plot.pdf}
    }
    \subfloat[]{
        \includegraphics[width=0.33\textwidth]{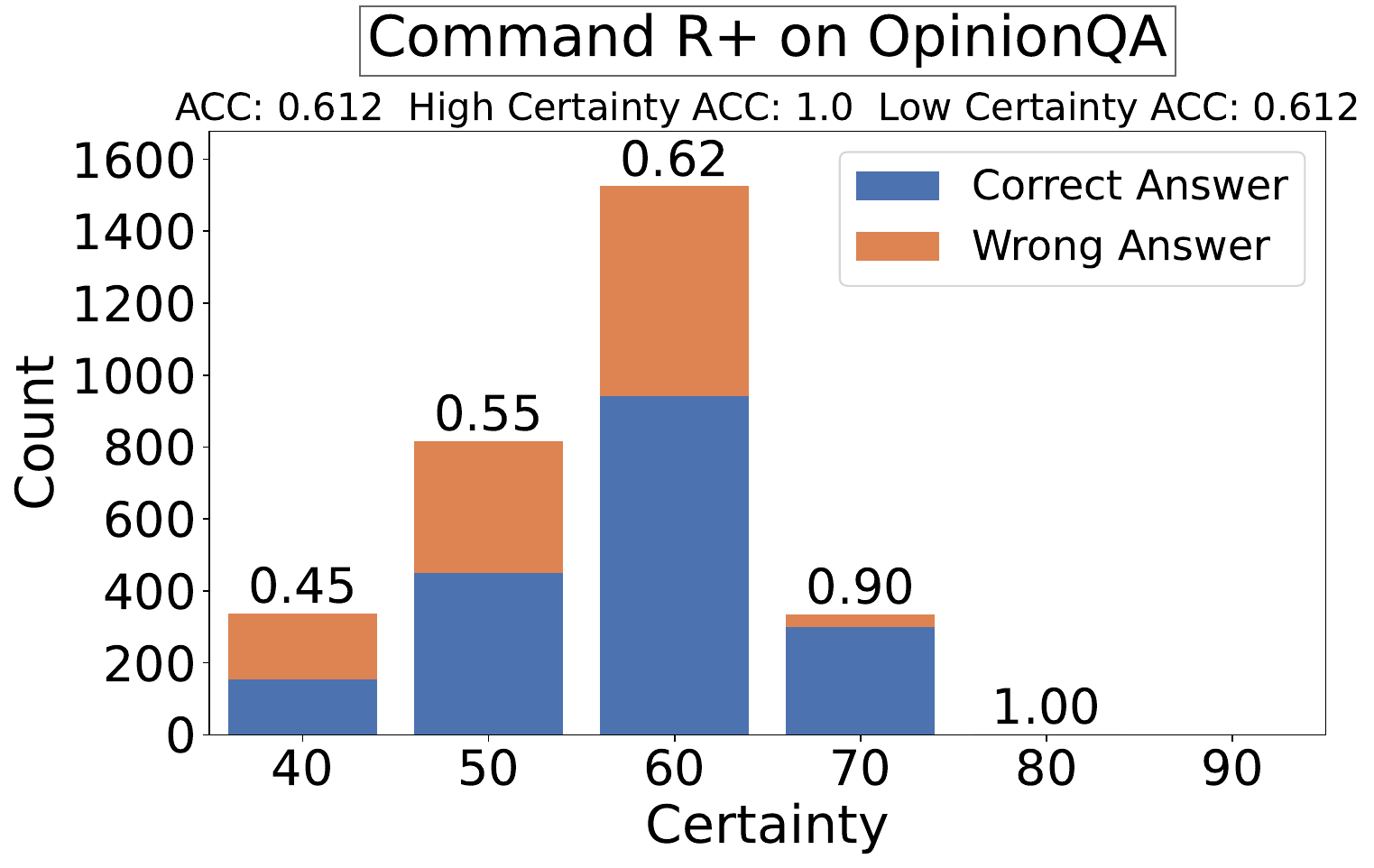}
    }
    \subfloat[]{
        \includegraphics[width=0.33\textwidth]{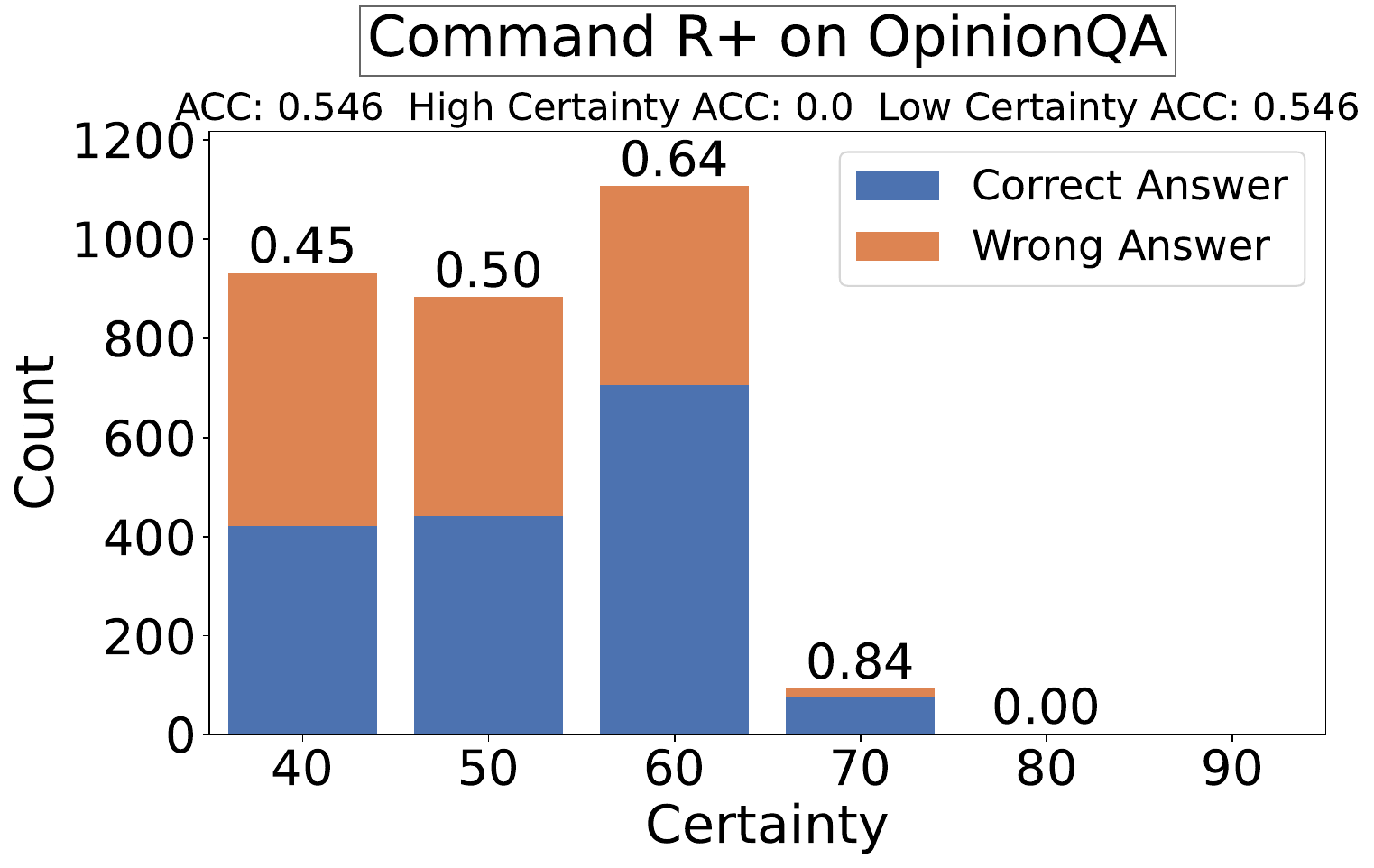}
    } 
\caption{\textbf{Distribution of LLM verbal certainty score when provided with different number of features.} The left column figures (a,d,g,j) show results when all features are provided. The middle column figures (b, e, h, k) show results when 3 features are provided. The right column figures (c, f, i, l) show results when only one feature is provided. The plots show the certainty distribution and corresponding accuracy of correct (blue) and wrong (orange) answers for Command R+ and GPT-4 models on PRISM and OpinionQA dataset. Each plot provides overall accuracy (ACC), high certainty accuracy (High Certainty ACC), and low certainty accuracy (Low Certainty ACC). The top of each bar shows the accuracy within that certainty bin. The certainty levels are truncated to be between 40 and 90 by adjusting values outside this range.}
\label{certainty_different_features}
\end{figure*}

In Table \ref{certainty_different_features}, we show the effect of using different number of persona variables on the certainty distribution. We observe that, on OpinionQA, GPT-4 and Command R+ show clear drop in confidence when fewer persona variables. On PRISM, since the quality difference is so large that the preference can be inferred regardless of the persona, only minimal change occurred to the certainty distribution.

\subsection{Prompts for LLM-as-a-Personalized-Judge}
\label{prompts}
In Figure \ref{prompt_tie} and Figure \ref{prompt_wo_tie}, we include the prompts that we used for PRISM. For other datasets, minor modifications are made to the prompt to fit the dataset.
\FloatBarrier
\begin{figure*}[htbp]
\begin{tcolorbox}[colback=blue!5!white, colframe=blue!75!black, title={Prompt for PRISM (without tie)}, width=\textwidth, boxrule=0.5mm]
Given the user profile provided below, select the response from AI assistant A or B that the user would most likely prefer. Declare your choice by using the format: "\textbf{[[A]]}" if you believe assistant A's response is more suitable, or "\textbf{[[B]]}" if assistant B's response is better suited. Additionally, assess your confidence in this decision by assigning a certainty level from 1 to 100. Use the following guidelines to assign the certainty level: \\

1--20 (\textbf{Uncertain}): The user profile provides insufficient or minimal evidence. The decision is largely based on weak or indirect hints.

21--40 (\textbf{Moderately Confident}): There is noticeable evidence supporting a preference, though it is not comprehensive, and other interpretations are possible.

41--60 (\textbf{Quite Confident}): You find clear and convincing evidence that supports your prediction, though it is not entirely decisive.

61--80 (\textbf{Confident}): The user profile contains strong evidence that clearly supports your prediction, with very little ambiguity.

81--100 (\textbf{Highly Confident}): The user profile provides direct and explicit evidence that decisively supports your prediction.

Ensure you enclose your chosen certainty level in double brackets, like so: \textbf{[[X]]}. \\

\textbf{[User Profile]} \\
\{user\_info\} \\

\textbf{[User Question]} \\
\{question\} \\

\textbf{[The Start of Assistant A's Answer]} \\
\{asst\_A\} \\
\textbf{[The End of Assistant A's Answer]} \\

\textbf{[The Start of Assistant B's Answer]} \\
\{asst\_B\} \\
\textbf{[The End of Assistant B's Answer]} \\

\textbf{[Answer]} \\
\textbf{[[}
\end{tcolorbox}
\caption{Prompt used for LLM-as-a-Personalized-Judge on PRISM. The placeholders \{user\_info\}, \{question\}, \{asst\_A\}, and \{asst\_B\} are replaced with the corresponding text from PRISM when querying the LLM. For other datasets, including EC, PR, and OpinionQA, minor modifications are made to the prompt to adapt to the specific characteristics of each dataset.}
\label{prompt_wo_tie}
\end{figure*}

\begin{figure*}[ht]
\begin{tcolorbox}[colback=blue!5!white, colframe=blue!75!black, title={Prompt for PRISM (with tie)}, width=\textwidth, boxrule=0.5mm]
Given the user profile provided below, select the response from AI assistant A or B that the user would most likely prefer. Declare your choice by using the format: "\textbf{[[A]]}" if you believe assistant A's response is more suitable, "\textbf{[[B]]}" if assistant B's response is better suited, or "\textbf{[[C]]}" for a tie. \\

\textbf{[User Profile]} \\
\{user\_info\} \\

\textbf{[User Question]} \\
\{question\} \\

\textbf{[The Start of Assistant A's Answer]} \\
\{asst\_A\} \\
\textbf{[The End of Assistant A's Answer]} \\

\textbf{[The Start of Assistant B's Answer]} \\
\{asst\_B\} \\
\textbf{[The End of Assistant B's Answer]} \\

\textbf{[Answer]} \\
\textbf{[[}
\end{tcolorbox}
\caption{Prompt used for LLM-as-a-Personalized-Judge on PRISM (with tie). The placeholders \{user\_info\}, \{question\}, \{asst\_A\}, and \{asst\_B\} are replaced with the corresponding text from PRISM when querying the LLM. For other datasets, including EC, PR, and OpinionQA, minor modifications are made to the prompt to adapt to the specific requirements of each dataset.}
\label{prompt_tie}
\end{figure*}

\end{document}